\pdfoutput=1

\documentclass[11pt]{article}

\usepackage{emnlp2021}

\usepackage{times}
\usepackage{latexsym}

\usepackage[T1]{fontenc}

\usepackage[utf8]{inputenc}

\usepackage{microtype}

\usepackage{graphicx}
\usepackage{booktabs}
\usepackage{amssymb}
\usepackage{amsmath}
\usepackage{longtable}
\usepackage{colortbl}
\usepackage{todonotes}

\usepackage{xspace}
\usepackage{amstext}
\usepackage{multirow}
\usepackage{tabularx}   
\usepackage{booktabs}   
\usepackage{arydshln}
\usepackage{subcaption}
\usepackage{colortbl}
\usepackage{xcolor}

\usepackage{breqn}   

\usepackage{pifont}

\title{What to Pre-Train on? \\ Efficient Intermediate Task Selection}

\author{Clifton Poth, Jonas Pfeiffer, Andreas R\"uckl\'e\thanks{~~Contributions made prior to joining Amazon.}, \and Iryna Gurevych \\
  Ubiquitous Knowledge Processing Lab, Technical University of Darmstadt \\
  \href{https://www.informatik.tu-darmstadt.de/ukp/ukp_home/index.en.jsp}{\texttt{www.ukp.tu-darmstadt.de}}}

\begin{document}
\maketitle
\begin{abstract} 

Intermediate task fine-tuning has been shown to culminate in large transfer gains across many NLP tasks. With an abundance of candidate datasets as well as pre-trained language models, it has become infeasible to experiment with all combinations  to find the best transfer setting.
In this work, we provide a comprehensive comparison of different methods for efficiently identifying beneficial tasks for intermediate transfer learning. We focus on parameter and computationally efficient adapter settings, highlight different data-availability scenarios, and provide expense estimates for each method. 
We experiment with a diverse set of 42 intermediate and 11 target English \textit{classification}, \textit{multiple choice, question answering}, and \textit{sequence tagging}  tasks. Our results demonstrate that efficient embedding based methods, which rely solely on the respective datasets, outperform computational expensive few-shot fine-tuning approaches. Our best methods achieve an average \textit{Regret@3} of 1\%  across all target tasks, demonstrating that we are able to efficiently identify the best datasets for intermediate training. \footnote{Code released at \url{https://github.com/Adapter-Hub/efficient-task-transfer}.}

\end{abstract}

\section{Introduction}

Large pre-trained language models (LMs) are continuously pushing the state of the art across various NLP tasks.
The established procedure performs self-supervised pre-training on a large text corpus and subsequently fine-tunes the model on a specific target task  \citep{devlinBERTPretrainingDeep2019,liuRoBERTaRobustlyOptimized2019}.
The same procedure has also been applied to adapter-based training strategies, which achieve 
on-par task performance to full model fine-tuning while being considerably more parameter efficient \cite{houlsbyParameterEfficientTransferLearning2019} and 
faster to train~\citep{ruckleAdapterDropEfficiencyAdapters2020}.\footnote{Adapters are new weights at every layer of a pre-trained transformer model. To fine-tune a model on a downstream task, all pre-trained transformer weights are frozen and only the newly introduced adapter weights are trained.} Besides being more efficient, adapters are also highly modular, enabling a wider range of transfer learning techniques~\citep{ pfeifferMADX, pfeifferAdapterFusionNonDestructiveTask2020, Pfeiffer20UNKs, ustun2020udapter, Vidoni2020OrthogonalLA, Rust2021tokenizer, Ansell2021MADG}.

Extending upon the established two-step learning procedure,  
incorporating \textit{intermediate} stages of knowledge transfer can yield further gains for fully fine-tuned models. 
For instance, \citet{phangSentenceEncodersSTILTs2018} sequentially fine-tune a pre-trained language model on a compatible \textit{intermediate} task before \textit{target} task fine-tuning. 
It has been shown that this is most effective for low-resource target tasks,
however, not all task combinations are beneficial and many yield decreased performances \citep{phangSentenceEncodersSTILTs2018,wangCanYouTell2019,pruksachatkunIntermediateTaskTransferLearning2020}.
The abundance of diverse labeled datasets as well as the continuous development of new pre-trained LMs calls for methods that efficiently identify \textit{intermediate} dataset that benefit the target task.  

So far, it is unclear how adapter-based approaches behave with intermediate fine-tuning.
In the \textbf{first part of this work}, we thus establish that this setup results in similar gains for adapters, as has been shown for full model fine-tuning \cite{phangSentenceEncodersSTILTs2018, pruksachatkunIntermediateTaskTransferLearning2020, Gururangan20dontstop}.  
Focusing on a low-resource target task setup,
we find that only a subset of intermediate adapters yield positive gains, while others hurt the performance considerably (see Table~\ref{tab:transfer_results} and Figure~\ref{fig:violin}). Our results demonstrate that it is necessary to obtain
methods that efficiently identify beneficial intermediately trained adapters.

In the \textbf{second part}, we leverage 
the transfer results from part one to automatically rank and identify beneficial intermediate tasks. 
With the rise of large publicly accessible repositories for NLP models \citep{wolfTransformersStateoftheArtNatural2020,pfeifferAdapterHubFrameworkAdapting2020},   the chances  of finding pre-trained models that yield positive transfer gains are high. 
However, it is infeasible to brute-force the identification of the best intermediate task. 
Existing approaches have focused on beneficial task selection for multi-task learning \cite{bingelIdentifyingBeneficialTask2017}, full fine-tuning of intermediate and target transformer-based LMs for NLP tasks \cite{vuExploringPredictingTransferability2020}, adapter-based models for vision tasks \cite{puigcerverScalableTransferLearning2020} and unsupervised approaches for zero-shot transfer for community question answering  \cite{ruckleMultiCQAZeroShotTransfer2020}. Each of these works require different types of data, such as intermediate task data and/or intermediate model weights, which, depending on the scenario,  are potentially not accessible.\footnote{\citet{bingelIdentifyingBeneficialTask2017} and \citet{vuExploringPredictingTransferability2020} require access to both intermediate task data and models, \citet{puigcerverScalableTransferLearning2020} require access to only the intermediate model, and \citet{ruckleMultiCQAZeroShotTransfer2020} only   to the intermediate task data.}  

In this work we thus aim to address the \textbf{efficiency} aspect of transfer learning in NLP from multiple different angles, resulting in the following \textbf{contributions}: 
\textbf{1)}~We focus on adapter-based transfer learning which is considerably more parameter \cite{houlsbyParameterEfficientTransferLearning2019} and computationally efficient than full model fine-tuning \cite{ruckleAdapterDropEfficiencyAdapters2020}, while achieving on-par performance;
\textbf{2)}~We 
evaluate 
sequential fine-tuning of adapter-based approaches on a diverse set of 42 intermediate and 11 target tasks (i.e. classification, multiple choice, question answering, and sequence tagging);
\textbf{3)}~We identify the best intermediate task for transfer learning, \textit{without} the necessity of \textit{computational expensive, explicit training} on all potential candidates. We compare different selection techniques, consolidating  previously proposed and  new methods;
\textbf{4)}~We provide a thorough analysis of the different techniques, available data scenarios, and task-, and model types, thus presenting deeper insights into the best approach for each respective setting;
\textbf{5)}~We provide \textit{computational cost estimates}, enabling informed decision making for trade-offs between \textit{expense} and downstream task performance.

\section{Related Work}

\subsection{Transfer between tasks}

\citet{phangSentenceEncodersSTILTs2018}  
show that training on 
intermediate tasks results in performance gains for many target tasks.
Subsequent work further explores the effects 
on 
more diverse sets of 
tasks \citep{wangCanYouTell2019,talmorMultiQAEmpiricalInvestigation2019,liuLinguisticKnowledgeTransferability2019,sapSocialIQACommonsense2019,pruksachatkunIntermediateTaskTransferLearning2020,vuExploringPredictingTransferability2020}.  
\citet{wangCanYouTell2019}, \citet{yogatamaLearningEvaluatingGeneral2019}, and \citet{pruksachatkunIntermediateTaskTransferLearning2020}  emphasizes the risks of catastrophic forgetting and negative transfer results, finding that the success of sequential transfer varies largely when considering different intermediate tasks. 

While previous work has shown that intermediate task training improves the performance on the target task in full fine-tuning setups, we establish that the same holds true for adapter-based training. 

\subsection{Predicting Beneficial Transfer Sources}\label{sec:predicting_transfer}
Automatically selecting  intermediate tasks that yield  transfer gains is critical when considering the increasing availability of tasks and models.  

Proxy estimators have been proposed to evaluate the transferability of pre-trained models towards a target task.  
\citet{nguyenLEEPNewMeasure2020}, \citet{liRankingNeuralCheckpoints2020} and \citet{deshpandeLinearizedFrameworkNew2021}  
estimate the transferability between classification tasks 
by building
an empirical classifier from the source and target task label distribution. 
\citet{puigcerverScalableTransferLearning2020} experiment with multiple model selection methods, 
including kNN proxy models to estimate the target task performance. 
In a similar direction,
\citet{renggliWhichModelTransfer2020} study proxy models based on kNN and linear classifiers, 
finding that a hybrid approach combination of \textit{task-aware} and \textit{task-agnostic} strategies yields the best results.

\citet{bingelIdentifyingBeneficialTask2017} find that gradients of the learning curves correlate with multi-task learning success. 
\citet{zamirTaskonomyDisentanglingTask2018} build a taxonomy of vision tasks, 
giving insights into non-trivial transfer relations between tasks. 
Multiple works propose 
using
embeddings that capture statistics, features, or the domain of a dataset.
\citet{edwardsNeuralStatistician2017} leverage variational autoencoders \citep{kingmaAutoEncodingVariationalBayes2014} to encode all samples of a dataset. 
\citet{jomaaDataset2VecLearningDataset2019} 
train
a dataset meta-feature extractor 
that can successfully capture the domain of a dataset. 
\citet{vuExploringPredictingTransferability2020}  
encode each training example of a dataset by
averaging over BERT's representations of the last layer.  
\citet{ruckleMultiCQAZeroShotTransfer2020} capture domain similarity by embedding dataset examples using a sentence embedding model. 
\citet{achilleTask2VecTaskEmbedding2019} and \citet{vuExploringPredictingTransferability2020}  compute task embeddings 
based on the Fisher Information Matrix of a probe network.

While many different methods have been  proposed, there lacks a direct comparison among them. Additionally, previous work has only focus on BERT,  which we find to behave considerably different to other model types such as RoBERTa for some methods.  In this work we aim to consolidate all methods and experiment with newer model types to provide a more thorough perspective.

\section{Adapter-Based Sequential Transfer}\label{sec:transfer}

We  
present a large-scale study on \emph{adapter-based} sequential fine-tuning, finding that 
around half of the task combinations yield no positive gains. This demonstrates the importance of finding approaches that efficiently identify suitable intermediate tasks. 

\subsection{Tasks}

We select \textit{QA} tasks from the MultiQA repository \citep{talmorMultiQAEmpiricalInvestigation2019}
and sequence \textit{tagging} tasks from \citet{liuLinguisticKnowledgeTransferability2019}.
Most of our \textit{classification} tasks are available in the (Super)GLUE \citep{wangGLUEMultiTaskBenchmark2018,wangSuperGLUEStickierBenchmark2019} benchmarks.
We  experiment with \textit{multiple choice} commonsense reasoning tasks to cover a broader range of different types, and domains. 
In total, we experiment with 
53 tasks, divided into 42 intermediate and 11 target tasks.\footnote{The choice for our intermediate and target task split was  motivated by previous work \cite[][\textit{inter alia}]{sapSocialIQACommonsense2019,vuExploringPredictingTransferability2020}. For more details see Appendix~\ref{app:tasks}.}

\subsection{Experimental Setup}\label{subsec:transfer_setup}
 
We  experiment with
BERT-base \cite{devlinBERTPretrainingDeep2019} and RoBERTa-base  \citep{liuRoBERTaRobustlyOptimized2019}, training adapters with the configuration proposed by \citet{pfeifferAdapterFusionNonDestructiveTask2020}.
 We adopt the two-stage sequential fine-tuning setup of \citet{phangSentenceEncodersSTILTs2018}, splitting the tasks 
in two disjoint subsets $\mathcal{S}$ and $\mathcal{T}$,  denoted as intermediate and target tasks, respectively.
For each pair $(s, t)$ with  $s \in \mathcal{S}$ and $t \in \mathcal{T}$, we first  
train a randomly initialized adapter on $s$ (keeping the base model's parameters fixed). 
We then fine-tune the
trained adapter on 
$t$.\footnote{
For more details please refer to Appendix~\ref{app:training}.}

For target task fine-tuning, we simulate a \textbf{low-resource setup} by limiting the maximum number of training examples on $t$ to 1000.
This choice is motivated by the observation that smaller target tasks 
benefit the most from sequential fine-tuning 
while at the same time revealing the largest performance variances \citep{phangSentenceEncodersSTILTs2018,vuExploringPredictingTransferability2020}.
Low-resource setups, thus, reflect the most beneficial application setting for our transfer learning strategy and also allow us to more thoroughly study different transfer relations.

\begin{figure}[]
    \centering
    \includegraphics[width=\columnwidth,trim=1.0cm 0.9cm 1.9cm 2.1cm,clip]{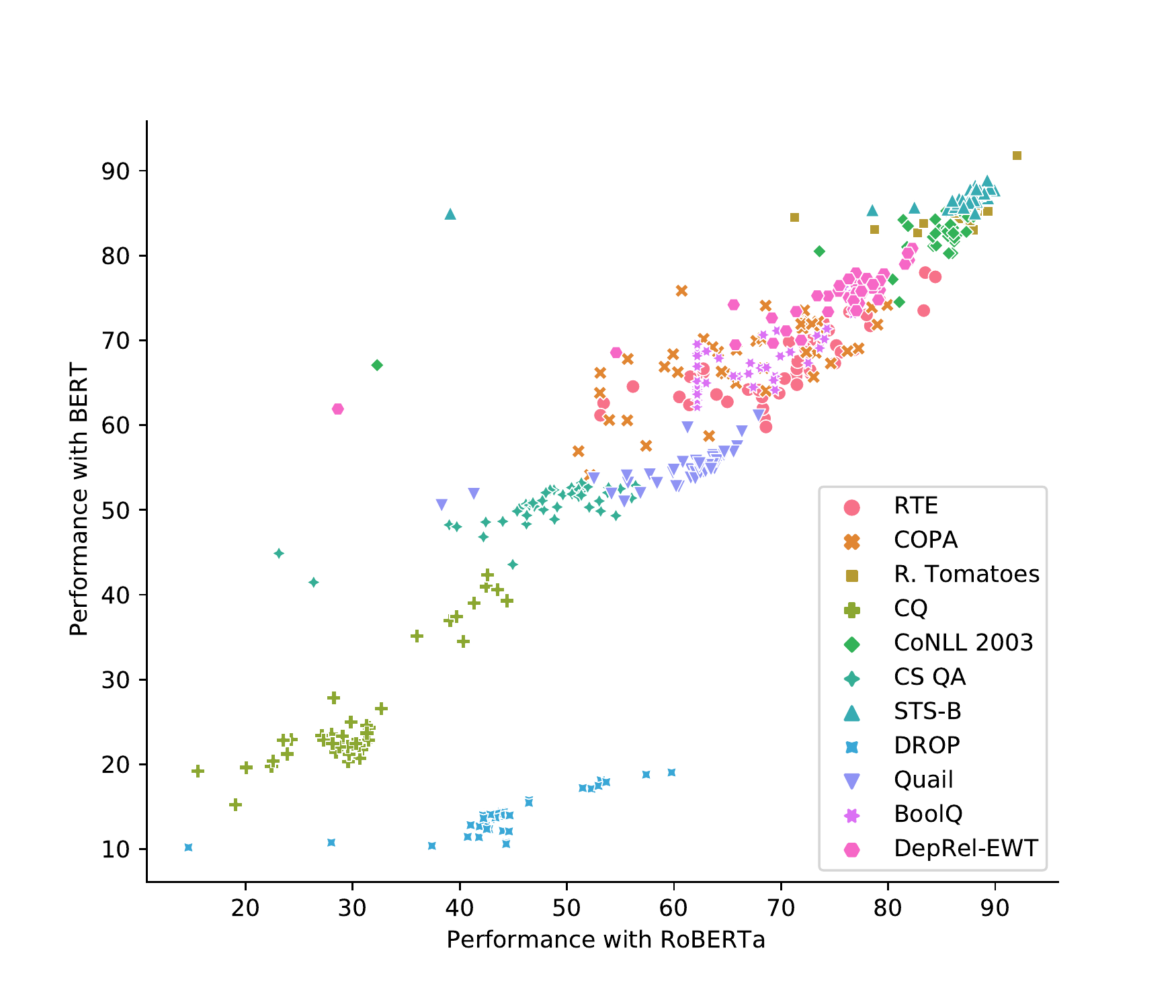}
    \caption{Comparison of transfer performance between BERT and RoBERTa for the respective target tasks, pre-trained on the 42 intermediate tasks.}
    \label{fig:corr}
    \vspace{-1.0em}
\end{figure}

\subsection{Results}

\begin{figure*}[t]
    \centering
    \includegraphics[width=\textwidth,trim=0.5cm 0.2cm 0.5cm 1cm,clip]{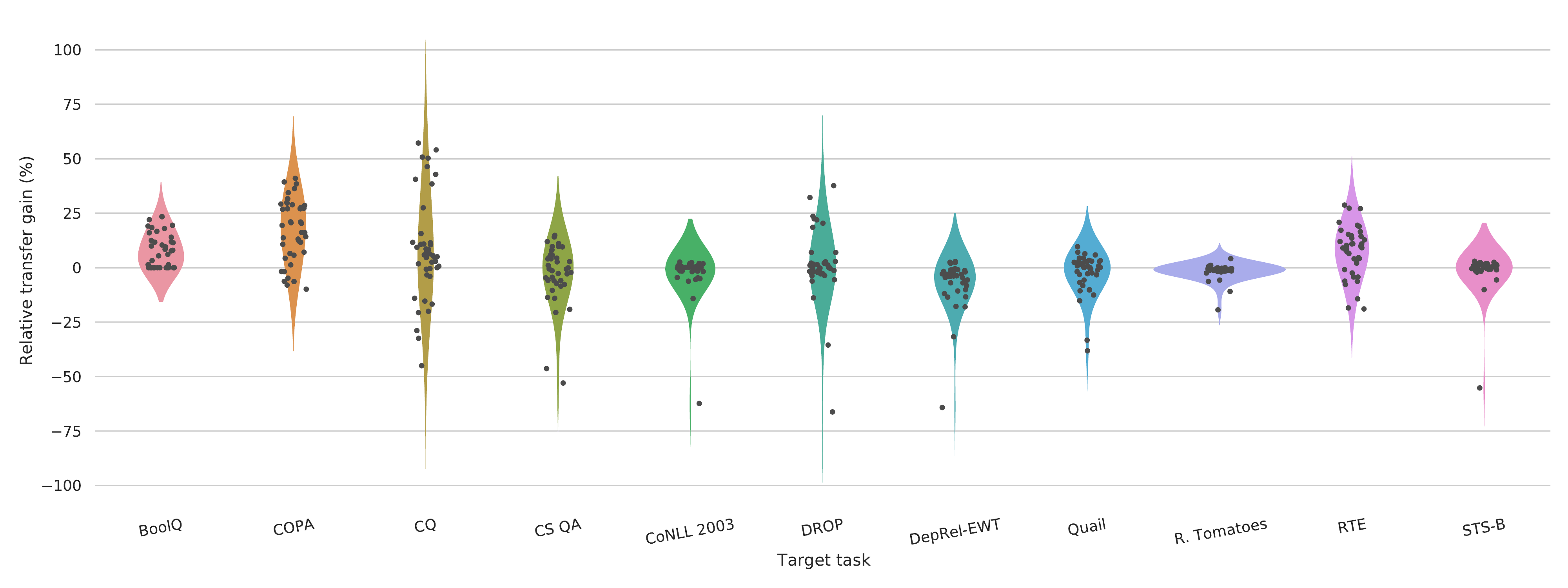}
     \vspace{-1em}
    \caption{Transfer gains/losses between all intermediate and target tasks with RoBERTa as base model.
    Each violin represents one target task where each dot represents the \emph{relative transfer gain} (y-axis) from one intermediate task.}
    \label{fig:violin}
\end{figure*}

Figure~\ref{fig:violin} shows the relative transfer gains and
Table~\ref{tab:transfer_results} lists the absolute scores of all intermediate and target task combinations for RoBERTa.\footnote{We list the corresponding transfer results for BERT in Table~\ref{tab:transfer_results_BERT} of the Appendix.}
We observe large variations in transfer gains (and losses) across the different combinations.  
Even though larger variances may be explained by a higher task difficulty (see `No Transfer' in Table~\ref{tab:transfer_results}), they also illustrate the heterogeneity and potential of sequential fine-tuning in our adapter-based setting. At the same time, we find several cases of transfer \emph{losses}---with up to 60\% lower performances (see Figure~\ref{fig:violin})---potentially occurring due to catastrophic forgetting. 
 
Overall,
for RoBERTa, 243 ($53\%$) transfer combinations yield positive transfer gains whereas 203 ($44\%$) yield losses. 
The mean of all transfer gains is $2.3\%$. However,
from our eleven target tasks \emph{only five} benefit on average (see `Avg. Transfer' in Table~\ref{tab:transfer_results}).
This illustrates the high risk of choosing the wrong intermediate tasks. 
Avoiding such hurtful combinations 
and efficiently identifying the best ones is necessary; evaluating all combinations is inefficient and often not feasible. 

We further find that the best performing intermediate tasks for BERT and RoBERTa overlap considerably as illustrated in Figure~\ref{fig:corr}, with transfer performances correlating with a Spearman correlation of 0.94 when averaged over all settings, and 0.68 when averaged per target task.  
 
\begin{table*}[t!]
    \centering
    \footnotesize
    \resizebox{\textwidth}{!}{
    \begin{tabular}{lp{1cm}p{1cm}p{1cm}p{1cm}p{1cm}p{1cm}p{1cm}p{1cm}p{1cm}p{1cm}p{1cm}}
    \toprule
    \bf Task & \bf BoolQ & \bf COPA & \bf CQ & \bf CS QA & \bf CoNLL 2003 & \bf DROP & \bf DepRel-EWT & \bf Quail & \bf R. Tomatoes & \bf RTE & \bf STS-B \\
    \midrule
    No Transfer & \textbf{62.17} & \textbf{56.68} & \textbf{28.24} & \textbf{49.12} & \textbf{85.74} & \textbf{43.42} & \textbf{79.96} & \textbf{61.94} & \textbf{88.35} & \textbf{65.56} & \textbf{87.35} \\
    \midrule
    Avg. Transfer & \textbf{66.93} & \textbf{66.49} & \textbf{30.60} & \textbf{47.88} & \textbf{83.80} & \textbf{43.96} & \textbf{74.59} & \textbf{60.50} & \textbf{86.95} & \textbf{70.38} & \textbf{86.31} \\
    \midrule
    ANLI & \cellcolor[rgb]{ .388,  .745,  .482}\textbf{76.75} & \cellcolor[rgb]{ .557,  .816,  .624}73.64 & \cellcolor[rgb]{ .835,  .933,  .859}32.67 & \cellcolor[rgb]{ .816,  .925,  .843}51.35 & \cellcolor[rgb]{ .98,  .992,  .984}85.82 & \cellcolor[rgb]{ .957,  .984,  .965}44.66 & \cellcolor[rgb]{ 1,  .984,  .937}76.77 & \cellcolor[rgb]{ .549,  .812,  .62}66.34 & \cellcolor[rgb]{ 1,  .988,  .957}87.64 & \cellcolor[rgb]{ .388,  .745,  .482}\textbf{84.40} & \cellcolor[rgb]{ .792,  .914,  .824}88.24 \\
    ART  & \cellcolor[rgb]{ .635,  .851,  .694}70.87 & \cellcolor[rgb]{ .416,  .757,  .506}79.00 & \cellcolor[rgb]{ 1,  .976,  .91}27.12 & \cellcolor[rgb]{ .675,  .867,  .725}53.02 & \cellcolor[rgb]{ 1,  .992,  .973}84.48 & \cellcolor[rgb]{ 1,  .976,  .914}41.01 & \cellcolor[rgb]{ 1,  .973,  .89}74.39 & \cellcolor[rgb]{ .788,  .914,  .824}64.00 & \cellcolor[rgb]{ 1,  .988,  .961}87.73 & \cellcolor[rgb]{ .745,  .894,  .784}73.43 & \cellcolor[rgb]{ .741,  .894,  .784}88.45 \\
    CoLA & \cellcolor[rgb]{ .969,  .988,  .973}63.01 & \cellcolor[rgb]{ .82,  .925,  .847}63.60 & \cellcolor[rgb]{ .886,  .953,  .906}31.27 & \cellcolor[rgb]{ .835,  .933,  .863}51.11 & \cellcolor[rgb]{ 1,  .996,  .996}85.53 & \cellcolor[rgb]{ 1,  .996,  .996}43.34 & \cellcolor[rgb]{ 1,  .996,  .984}79.24 & \cellcolor[rgb]{ .847,  .937,  .871}63.46 & \cellcolor[rgb]{ 1,  .996,  .984}88.14 & \cellcolor[rgb]{ .957,  .984,  .965}66.93 & \cellcolor[rgb]{ .98,  .992,  .984}87.45 \\
    CoNLL'00 & 62.17 & \cellcolor[rgb]{ .984,  .992,  .984}57.40 & \cellcolor[rgb]{ .922,  .969,  .937}30.32 & \cellcolor[rgb]{ 1,  .957,  .839}44.95 & \cellcolor[rgb]{ .573,  .824,  .639}87.30 & \cellcolor[rgb]{ .969,  .988,  .973}44.33 & \cellcolor[rgb]{ .573,  .824,  .639}81.58 & \cellcolor[rgb]{ 1,  .918,  .671}54.17 & \cellcolor[rgb]{ 1,  .98,  .929}87.19 & \cellcolor[rgb]{ 1,  .918,  .671}61.52 & \cellcolor[rgb]{ .82,  .925,  .847}88.11 \\
    Cosmos QA & \cellcolor[rgb]{ .494,  .788,  .573}74.28 & \cellcolor[rgb]{ .427,  .765,  .518}78.48 & \cellcolor[rgb]{ .941,  .976,  .953}29.83 & \cellcolor[rgb]{ .608,  .839,  .671}53.81 & \cellcolor[rgb]{ 1,  .992,  .969}84.17 & \cellcolor[rgb]{ .992,  1,  .996}43.66 & \cellcolor[rgb]{ 1,  .973,  .89}74.37 & \cellcolor[rgb]{ .631,  .847,  .686}65.56 & \cellcolor[rgb]{ 1,  .996,  .984}88.12 & \cellcolor[rgb]{ .561,  .816,  .627}79.21 & \cellcolor[rgb]{ .847,  .937,  .871}88.00 \\
    DuoRC-p & 62.17 & \cellcolor[rgb]{ .788,  .914,  .824}64.76 & \cellcolor[rgb]{ .592,  .831,  .655}39.11 & \cellcolor[rgb]{ .961,  .984,  .965}49.63 & \cellcolor[rgb]{ 1,  .996,  .992}85.42 & \cellcolor[rgb]{ .702,  .875,  .745}51.48 & \cellcolor[rgb]{ 1,  .98,  .925}76.33 & \cellcolor[rgb]{ 1,  .98,  .925}60.20 & \cellcolor[rgb]{ 1,  .976,  .918}86.98 & \cellcolor[rgb]{ .808,  .922,  .839}71.55 & \cellcolor[rgb]{ .722,  .886,  .765}88.54 \\
    DuoRC-s & 62.17 & \cellcolor[rgb]{ .714,  .882,  .757}67.68 & \cellcolor[rgb]{ .463,  .776,  .545}42.45 & \cellcolor[rgb]{ .82,  .925,  .847}51.29 & \cellcolor[rgb]{ 1,  .996,  .996}85.66 & \cellcolor[rgb]{ .671,  .863,  .722}52.29 & \cellcolor[rgb]{ 1,  .976,  .91}75.46 & 61.94 & \cellcolor[rgb]{ 1,  .976,  .914}86.92 & \cellcolor[rgb]{ .769,  .906,  .804}72.78 & \cellcolor[rgb]{ .729,  .886,  .769}88.51 \\
    EmoContext & \cellcolor[rgb]{ .729,  .89,  .773}68.64 & \cellcolor[rgb]{ .808,  .922,  .835}64.12 & \cellcolor[rgb]{ 1,  .906,  .627}23.52 & \cellcolor[rgb]{ 1,  .976,  .914}46.96 & \cellcolor[rgb]{ .827,  .929,  .855}86.37 & \cellcolor[rgb]{ 1,  .988,  .961}42.34 & \cellcolor[rgb]{ 1,  .988,  .953}77.69 & \cellcolor[rgb]{ .996,  1,  .996}62.01 & \cellcolor[rgb]{ 1,  .992,  .98}88.03 & \cellcolor[rgb]{ .808,  .922,  .839}71.48 & \cellcolor[rgb]{ 1,  .996,  .988}86.91 \\
    Emotion & \cellcolor[rgb]{ .859,  .941,  .882}65.54 & \cellcolor[rgb]{ .906,  .961,  .922}60.36 & \cellcolor[rgb]{ 1,  .886,  .553}22.58 & \cellcolor[rgb]{ 1,  .965,  .859}45.54 & \cellcolor[rgb]{ 1,  .992,  .973}84.41 & \cellcolor[rgb]{ 1,  .988,  .957}42.22 & \cellcolor[rgb]{ 1,  .984,  .941}77.03 & \cellcolor[rgb]{ 1,  .976,  .914}59.99 & \cellcolor[rgb]{ 1,  .992,  .976}88.01 & \cellcolor[rgb]{ 1,  .965,  .871}63.97 & \cellcolor[rgb]{ .945,  .976,  .953}87.59 \\
    GED-FCE & 62.17 & \cellcolor[rgb]{ 1,  .953,  .82}55.68 & \cellcolor[rgb]{ .933,  .973,  .945}30.05 & \cellcolor[rgb]{ 1,  .973,  .886}46.26 & \cellcolor[rgb]{ .878,  .949,  .898}86.19 & \cellcolor[rgb]{ 1,  .992,  .969}42.52 & \cellcolor[rgb]{ 1,  .996,  .992}79.61 & \cellcolor[rgb]{ 1,  .976,  .914}59.92 & \cellcolor[rgb]{ 1,  .988,  .957}87.64 & \cellcolor[rgb]{ 1,  .898,  .592}60.51 & \cellcolor[rgb]{ 1,  .992,  .98}86.49 \\
    Hellaswag & \cellcolor[rgb]{ .698,  .875,  .745}69.44 & \cellcolor[rgb]{ .459,  .776,  .545}77.24 & \cellcolor[rgb]{ .91,  .965,  .925}30.67 & \cellcolor[rgb]{ .42,  .761,  .51}56.05 & \cellcolor[rgb]{ .965,  .988,  .973}85.87 & \cellcolor[rgb]{ 1,  .992,  .98}42.86 & \cellcolor[rgb]{ 1,  .976,  .91}75.34 & \cellcolor[rgb]{ .988,  .996,  .992}62.06 & \cellcolor[rgb]{ 1,  .988,  .957}87.64 & \cellcolor[rgb]{ .635,  .851,  .694}76.82 & \cellcolor[rgb]{ .69,  .871,  .737}88.67 \\
    HotpotQA & 62.17 & \cellcolor[rgb]{ .937,  .976,  .949}59.12 & \cellcolor[rgb]{ .459,  .776,  .541}42.58 & \cellcolor[rgb]{ 1,  .91,  .635}39.72 & \cellcolor[rgb]{ 1,  .941,  .773}73.59 & \cellcolor[rgb]{ .62,  .843,  .678}53.68 & \cellcolor[rgb]{ 1,  .875,  .506}54.60 & \cellcolor[rgb]{ 1,  .929,  .722}55.36 & \cellcolor[rgb]{ 1,  .925,  .702}83.30 & \cellcolor[rgb]{ 1,  .941,  .773}62.74 & \cellcolor[rgb]{ 1,  .992,  .976}86.26 \\
    IMDb & \cellcolor[rgb]{ .753,  .898,  .792}68.09 & \cellcolor[rgb]{ .843,  .933,  .867}62.76 & \cellcolor[rgb]{ 1,  .992,  .98}28.04 & \cellcolor[rgb]{ 1,  .969,  .886}46.18 & \cellcolor[rgb]{ .957,  .984,  .965}85.90 & \cellcolor[rgb]{ 1,  .996,  .992}43.30 & \cellcolor[rgb]{ 1,  .988,  .953}77.55 & \cellcolor[rgb]{ 1,  .996,  .984}61.60 & \cellcolor[rgb]{ .898,  .961,  .914}88.97 & \cellcolor[rgb]{ .91,  .965,  .922}68.45 & \cellcolor[rgb]{ 1,  .992,  .976}86.38 \\
    MIT Movie & 62.17 & \cellcolor[rgb]{ 1,  .839,  .361}53.12 & \cellcolor[rgb]{ .886,  .953,  .902}31.32 & \cellcolor[rgb]{ 1,  .949,  .8}44.01 & \cellcolor[rgb]{ .537,  .808,  .608}87.43 & \cellcolor[rgb]{ .957,  .984,  .965}44.59 & \cellcolor[rgb]{ 1,  .992,  .973}78.58 & \cellcolor[rgb]{ 1,  .961,  .851}58.42 & \cellcolor[rgb]{ 1,  .988,  .953}87.56 & \cellcolor[rgb]{ .925,  .969,  .937}67.94 & \cellcolor[rgb]{ 1,  .996,  .992}87.05 \\
    MNLI & \cellcolor[rgb]{ .427,  .761,  .518}75.86 & \cellcolor[rgb]{ .486,  .788,  .569}76.20 & \cellcolor[rgb]{ .878,  .949,  .898}31.52 & \cellcolor[rgb]{ 1,  .996,  .984}48.80 & \cellcolor[rgb]{ 1,  .996,  .996}85.59 & \cellcolor[rgb]{ .98,  .992,  .984}43.95 & \cellcolor[rgb]{ 1,  .957,  .831}71.41 & \cellcolor[rgb]{ 1,  .992,  .969}61.27 & \cellcolor[rgb]{ 1,  .996,  .996}88.31 & \cellcolor[rgb]{ .42,  .761,  .51}83.47 & \cellcolor[rgb]{ .553,  .816,  .62}89.25 \\
    MRPC & \cellcolor[rgb]{ .918,  .965,  .929}64.19 & \cellcolor[rgb]{ .69,  .871,  .737}68.56 & \cellcolor[rgb]{ .973,  .988,  .976}29.06 & \cellcolor[rgb]{ 1,  .992,  .973}48.44 & \cellcolor[rgb]{ .882,  .953,  .898}86.18 & 43.43 & \cellcolor[rgb]{ 1,  .984,  .941}77.05 & \cellcolor[rgb]{ .898,  .957,  .914}62.95 & \cellcolor[rgb]{ 1,  .984,  .941}87.35 & \cellcolor[rgb]{ .769,  .906,  .804}72.71 & \cellcolor[rgb]{ .616,  .843,  .678}88.97 \\
    MultiRC & \cellcolor[rgb]{ .502,  .796,  .58}74.05 & \cellcolor[rgb]{ .6,  .835,  .663}71.88 & \cellcolor[rgb]{ .914,  .965,  .925}30.59 & \cellcolor[rgb]{ .816,  .925,  .843}51.35 & \cellcolor[rgb]{ 1,  .98,  .925}81.86 & \cellcolor[rgb]{ .984,  .996,  .988}43.89 & \cellcolor[rgb]{ 1,  .965,  .871}73.38 & \cellcolor[rgb]{ .718,  .882,  .761}64.70 & \cellcolor[rgb]{ 1,  .98,  .929}87.17 & \cellcolor[rgb]{ .6,  .835,  .663}77.98 & \cellcolor[rgb]{ .835,  .933,  .859}88.06 \\
    NewsQA & 62.17 & \cellcolor[rgb]{ .698,  .875,  .745}68.24 & \cellcolor[rgb]{ .424,  .761,  .514}43.52 & \cellcolor[rgb]{ 1,  .996,  .996}49.09 & \cellcolor[rgb]{ 1,  .976,  .918}81.41 & \cellcolor[rgb]{ .639,  .851,  .694}53.16 & \cellcolor[rgb]{ 1,  .929,  .718}65.57 & \cellcolor[rgb]{ 1,  .984,  .933}60.44 & \cellcolor[rgb]{ 1,  .98,  .933}87.22 & \cellcolor[rgb]{ .792,  .914,  .824}72.06 & \cellcolor[rgb]{ .929,  .973,  .941}87.66 \\
    POS-Co.'03 & 62.17 & \cellcolor[rgb]{ 1,  .753,  0}51.08 & \cellcolor[rgb]{ 1,  .839,  .357}20.09 & \cellcolor[rgb]{ 1,  .753,  0}23.11 & \cellcolor[rgb]{ .616,  .839,  .675}87.14 & \cellcolor[rgb]{ 1,  .976,  .906}40.74 & \cellcolor[rgb]{ .502,  .792,  .58}81.84 & \cellcolor[rgb]{ 1,  .784,  .125}41.31 & \cellcolor[rgb]{ 1,  .753,  0}71.24 & \cellcolor[rgb]{ 1,  .753,  0}53.14 & \cellcolor[rgb]{ 1,  .953,  .816}78.53 \\
    POS-EWT & 62.17 & \cellcolor[rgb]{ 1,  .878,  .514}53.96 & \cellcolor[rgb]{ .89,  .957,  .906}31.21 & \cellcolor[rgb]{ 1,  .969,  .886}46.22 & \cellcolor[rgb]{ .443,  .769,  .529}87.76 & \cellcolor[rgb]{ .969,  .988,  .973}44.31 & \cellcolor[rgb]{ .388,  .745,  .482}\textbf{82.25} & \cellcolor[rgb]{ 1,  .933,  .733}55.71 & \cellcolor[rgb]{ 1,  .976,  .91}86.81 & \cellcolor[rgb]{ 1,  .812,  .243}56.17 & \cellcolor[rgb]{ .953,  .98,  .961}87.55 \\
    QNLI & \cellcolor[rgb]{ .522,  .8,  .596}73.61 & \cellcolor[rgb]{ .6,  .835,  .659}72.00 & \cellcolor[rgb]{ .706,  .878,  .753}36.01 & \cellcolor[rgb]{ .765,  .902,  .8}51.94 & \cellcolor[rgb]{ .941,  .976,  .949}85.96 & \cellcolor[rgb]{ .886,  .953,  .906}46.47 & \cellcolor[rgb]{ 1,  .984,  .937}76.91 & \cellcolor[rgb]{ .722,  .886,  .765}64.67 & \cellcolor[rgb]{ 1,  .988,  .957}87.64 & \cellcolor[rgb]{ .831,  .933,  .859}70.76 & \cellcolor[rgb]{ .596,  .831,  .659}89.07 \\
    QQP  & \cellcolor[rgb]{ .965,  .988,  .973}63.03 & \cellcolor[rgb]{ .596,  .831,  .659}72.16 & \cellcolor[rgb]{ 1,  .922,  .686}24.27 & \cellcolor[rgb]{ 1,  .996,  .988}48.85 & \cellcolor[rgb]{ 1,  .98,  .925}81.77 & \cellcolor[rgb]{ 1,  .984,  .941}41.79 & \cellcolor[rgb]{ 1,  .945,  .788}69.14 & \cellcolor[rgb]{ 1,  .996,  .984}61.57 & \cellcolor[rgb]{ 1,  .988,  .953}87.58 & \cellcolor[rgb]{ .769,  .906,  .804}72.71 & \cellcolor[rgb]{ .49,  .788,  .569}89.51 \\
    QuaRTz & \cellcolor[rgb]{ .698,  .875,  .745}69.44 & \cellcolor[rgb]{ .686,  .871,  .737}68.60 & \cellcolor[rgb]{ 1,  .98,  .925}27.30 & \cellcolor[rgb]{ .886,  .953,  .906}50.48 & \cellcolor[rgb]{ .929,  .973,  .941}86.01 & \cellcolor[rgb]{ 1,  .992,  .973}42.69 & \cellcolor[rgb]{ 1,  .992,  .973}78.56 & \cellcolor[rgb]{ .949,  .98,  .957}62.47 & \cellcolor[rgb]{ 1,  .996,  .988}88.16 & \cellcolor[rgb]{ .914,  .965,  .925}68.30 & \cellcolor[rgb]{ .784,  .91,  .82}88.27 \\
    Quoref & 62.17 & \cellcolor[rgb]{ .894,  .957,  .914}60.72 & \cellcolor[rgb]{ .545,  .812,  .616}40.34 & \cellcolor[rgb]{ 1,  .976,  .91}46.83 & \cellcolor[rgb]{ .969,  .988,  .973}85.86 & \cellcolor[rgb]{ .647,  .855,  .702}52.95 & \cellcolor[rgb]{ 1,  .984,  .941}76.98 & \cellcolor[rgb]{ .918,  .969,  .929}62.75 & \cellcolor[rgb]{ 1,  .98,  .922}87.04 & \cellcolor[rgb]{ .808,  .922,  .839}71.48 & \cellcolor[rgb]{ .804,  .918,  .835}88.19 \\
    RACE & \cellcolor[rgb]{ .392,  .749,  .486}76.72 & \cellcolor[rgb]{ .596,  .835,  .659}72.04 & \cellcolor[rgb]{ 1,  .914,  .659}23.91 & \cellcolor[rgb]{ .663,  .863,  .714}53.15 & \cellcolor[rgb]{ 1,  .976,  .91}81.05 & \cellcolor[rgb]{ 1,  .988,  .965}42.46 & \cellcolor[rgb]{ 1,  .929,  .722}65.74 & \cellcolor[rgb]{ .388,  .745,  .482}\textbf{67.89} & \cellcolor[rgb]{ 1,  .98,  .929}87.19 & \cellcolor[rgb]{ .424,  .761,  .514}83.32 & \cellcolor[rgb]{ .816,  .925,  .843}88.13 \\
    ReCoRD & 62.17 & \cellcolor[rgb]{ .827,  .929,  .855}63.28 & 28.25 & \cellcolor[rgb]{ 1,  .973,  .898}46.50 & \cellcolor[rgb]{ 1,  .992,  .969}84.21 & \cellcolor[rgb]{ 1,  .984,  .945}41.85 & \cellcolor[rgb]{ 1,  .961,  .839}71.87 & \cellcolor[rgb]{ .827,  .929,  .855}63.63 & \cellcolor[rgb]{ 1,  .976,  .918}86.96 & \cellcolor[rgb]{ .812,  .922,  .843}71.41 & \cellcolor[rgb]{ 1,  .996,  .992}86.98 \\
    SICK & \cellcolor[rgb]{ .569,  .82,  .635}72.53 & \cellcolor[rgb]{ .561,  .82,  .627}73.48 & \cellcolor[rgb]{ .949,  .98,  .957}29.66 & \cellcolor[rgb]{ .6,  .835,  .663}53.89 & \cellcolor[rgb]{ 1,  .996,  .992}85.45 & \cellcolor[rgb]{ .976,  .992,  .98}44.11 & \cellcolor[rgb]{ 1,  .996,  .984}79.23 & \cellcolor[rgb]{ .827,  .929,  .855}63.63 & \cellcolor[rgb]{ 1,  .992,  .973}87.94 & \cellcolor[rgb]{ .694,  .875,  .741}75.02 & \cellcolor[rgb]{ .784,  .914,  .82}88.26 \\
    SNLI & \cellcolor[rgb]{ .69,  .871,  .737}69.61 & \cellcolor[rgb]{ .588,  .831,  .651}72.36 & \cellcolor[rgb]{ 1,  .82,  .278}19.08 & \cellcolor[rgb]{ 1,  .933,  .733}42.21 & \cellcolor[rgb]{ 1,  .753,  0}32.29 & \cellcolor[rgb]{ 1,  .753,  0}14.66 & \cellcolor[rgb]{ 1,  .753,  0}28.62 & \cellcolor[rgb]{ 1,  .898,  .6}52.53 & \cellcolor[rgb]{ 1,  .918,  .671}82.78 & \cellcolor[rgb]{ .651,  .855,  .706}76.39 & \cellcolor[rgb]{ 1,  .992,  .969}85.90 \\
    SQuAD & 62.17 & \cellcolor[rgb]{ .694,  .875,  .741}68.36 & \cellcolor[rgb]{ .569,  .82,  .635}39.71 & \cellcolor[rgb]{ .835,  .933,  .863}51.09 & \cellcolor[rgb]{ .8,  .918,  .831}86.48 & \cellcolor[rgb]{ .478,  .784,  .561}57.40 & \cellcolor[rgb]{ 1,  .98,  .925}76.33 & \cellcolor[rgb]{ .996,  1,  .996}61.99 & \cellcolor[rgb]{ 1,  .969,  .875}86.21 & \cellcolor[rgb]{ .784,  .91,  .816}72.27 & \cellcolor[rgb]{ .89,  .957,  .906}87.82 \\
    SQuAD 2.0 & \cellcolor[rgb]{ .741,  .894,  .784}68.34 & \cellcolor[rgb]{ .565,  .82,  .631}73.24 & \cellcolor[rgb]{ .388,  .745,  .482}\textbf{44.40} & \cellcolor[rgb]{ .89,  .957,  .91}50.43 & \cellcolor[rgb]{ .89,  .957,  .91}86.14 & \cellcolor[rgb]{ .388,  .745,  .482}\textbf{59.78} & \cellcolor[rgb]{ 1,  .984,  .941}76.99 & \cellcolor[rgb]{ .812,  .922,  .839}63.79 & \cellcolor[rgb]{ 1,  .992,  .973}87.90 & \cellcolor[rgb]{ .675,  .867,  .725}75.60 & \cellcolor[rgb]{ .596,  .835,  .659}89.06 \\
    SST-2 & \cellcolor[rgb]{ .792,  .914,  .827}67.14 & \cellcolor[rgb]{ .761,  .902,  .796}65.84 & \cellcolor[rgb]{ .996,  1,  .996}28.43 & \cellcolor[rgb]{ 1,  .984,  .949}47.81 & \cellcolor[rgb]{ 1,  .996,  .996}85.66 & \cellcolor[rgb]{ 1,  .992,  .973}42.73 & \cellcolor[rgb]{ 1,  .984,  .945}77.28 & \cellcolor[rgb]{ 1,  .988,  .949}60.82 & \cellcolor[rgb]{ .388,  .745,  .482}\textbf{92.03} & \cellcolor[rgb]{ .863,  .945,  .886}69.82 & \cellcolor[rgb]{ 1,  .996,  .984}86.75 \\
    ST-PMB & 62.17 & \cellcolor[rgb]{ 1,  .8,  .192}52.16 & \cellcolor[rgb]{ 1,  .753,  0}15.53 & \cellcolor[rgb]{ 1,  .78,  .122}26.36 & \cellcolor[rgb]{ .525,  .804,  .6}87.47 & \cellcolor[rgb]{ 1,  .867,  .463}28.00 & \cellcolor[rgb]{ .475,  .78,  .553}81.94 & \cellcolor[rgb]{ 1,  .753,  0}38.31 & \cellcolor[rgb]{ 1,  .859,  .435}78.71 & \cellcolor[rgb]{ 1,  .757,  .02}53.43 & \cellcolor[rgb]{ 1,  .753,  0}39.12 \\
    SWAG & \cellcolor[rgb]{ .702,  .878,  .749}69.30 & \cellcolor[rgb]{ .529,  .804,  .6}74.64 & \cellcolor[rgb]{ .976,  .992,  .98}28.96 & \cellcolor[rgb]{ .51,  .796,  .584}55.00 & \cellcolor[rgb]{ 1,  .996,  .996}85.61 & \cellcolor[rgb]{ 1,  .996,  .992}43.30 & \cellcolor[rgb]{ 1,  .984,  .941}76.97 & \cellcolor[rgb]{ .82,  .925,  .847}63.71 & \cellcolor[rgb]{ 1,  .996,  .996}88.35 & \cellcolor[rgb]{ .714,  .882,  .757}74.44 & \cellcolor[rgb]{ .537,  .808,  .608}89.32 \\
    SciCite & \cellcolor[rgb]{ .843,  .937,  .867}65.98 & \cellcolor[rgb]{ .796,  .918,  .827}64.44 & \cellcolor[rgb]{ .984,  .992,  .984}28.75 & \cellcolor[rgb]{ 1,  .984,  .945}47.71 & \cellcolor[rgb]{ .91,  .965,  .925}86.07 & \cellcolor[rgb]{ 1,  .988,  .953}42.18 & \cellcolor[rgb]{ 1,  .992,  .976}78.95 & \cellcolor[rgb]{ .953,  .98,  .961}62.40 & \cellcolor[rgb]{ 1,  .988,  .965}87.77 & \cellcolor[rgb]{ .902,  .961,  .918}68.59 & \cellcolor[rgb]{ 1,  .996,  .984}86.63 \\
    SciTail & \cellcolor[rgb]{ .584,  .827,  .647}72.13 & \cellcolor[rgb]{ .6,  .835,  .659}72.00 & \cellcolor[rgb]{ .937,  .976,  .949}29.91 & \cellcolor[rgb]{ .6,  .835,  .663}53.89 & \cellcolor[rgb]{ 1,  .992,  .98}84.74 & \cellcolor[rgb]{ .988,  .996,  .992}43.74 & \cellcolor[rgb]{ 1,  .988,  .953}77.72 & \cellcolor[rgb]{ .816,  .925,  .843}63.77 & \cellcolor[rgb]{ 1,  .992,  .969}87.82 & \cellcolor[rgb]{ .729,  .89,  .773}73.94 & \cellcolor[rgb]{ .388,  .745,  .482}\textbf{89.93} \\
    Social IQA & \cellcolor[rgb]{ .533,  .808,  .604}73.36 & \cellcolor[rgb]{ .388,  .745,  .482}\textbf{79.92} & \cellcolor[rgb]{ .902,  .961,  .918}30.88 & \cellcolor[rgb]{ .388,  .745,  .482}\textbf{56.41} & \cellcolor[rgb]{ .929,  .973,  .941}86.00 & \cellcolor[rgb]{ 1,  .996,  .996}43.34 & \cellcolor[rgb]{ 1,  .992,  .969}78.38 & \cellcolor[rgb]{ .596,  .831,  .659}65.91 & \cellcolor[rgb]{ 1,  .996,  .992}88.27 & \cellcolor[rgb]{ .588,  .827,  .651}78.34 & \cellcolor[rgb]{ .714,  .882,  .757}88.57 \\
    TREC & \cellcolor[rgb]{ .78,  .91,  .816}67.41 & \cellcolor[rgb]{ .918,  .965,  .929}59.92 & \cellcolor[rgb]{ .878,  .953,  .898}31.47 & \cellcolor[rgb]{ 1,  .988,  .957}48.04 & \cellcolor[rgb]{ .906,  .961,  .918}86.09 & \cellcolor[rgb]{ 1,  .992,  .98}42.86 & \cellcolor[rgb]{ 1,  .988,  .961}78.02 & \cellcolor[rgb]{ .957,  .984,  .965}62.38 & \cellcolor[rgb]{ 1,  .996,  .984}88.14 & \cellcolor[rgb]{ .847,  .937,  .871}70.32 & \cellcolor[rgb]{ 1,  .996,  .988}86.94 \\
    WNUT17 & 62.17 & \cellcolor[rgb]{ 1,  .839,  .357}53.08 & \cellcolor[rgb]{ 1,  .996,  .988}28.13 & \cellcolor[rgb]{ 1,  .961,  .855}45.36 & \cellcolor[rgb]{ .388,  .745,  .482}\textbf{87.96} & \cellcolor[rgb]{ .98,  .992,  .984}44.03 & \cellcolor[rgb]{ 1,  .984,  .949}77.51 & \cellcolor[rgb]{ 1,  .933,  .729}55.59 & \cellcolor[rgb]{ 1,  .988,  .949}87.54 & \cellcolor[rgb]{ 1,  .918,  .667}61.44 & \cellcolor[rgb]{ 1,  .992,  .969}86.00 \\
    WiC  & 62.17 & \cellcolor[rgb]{ .576,  .824,  .639}72.88 & \cellcolor[rgb]{ .953,  .98,  .961}29.55 & \cellcolor[rgb]{ .753,  .898,  .792}52.09 & \cellcolor[rgb]{ 1,  .996,  .996}85.73 & \cellcolor[rgb]{ 1,  .992,  .973}42.74 & \cellcolor[rgb]{ 1,  .992,  .98}79.08 & \cellcolor[rgb]{ .878,  .949,  .894}63.16 & \cellcolor[rgb]{ 1,  .992,  .973}87.92 & \cellcolor[rgb]{ .914,  .965,  .929}68.23 & \cellcolor[rgb]{ .847,  .937,  .871}88.00 \\
    WikiHop & 62.18 & \cellcolor[rgb]{ 1,  .953,  .812}55.64 & \cellcolor[rgb]{ .506,  .796,  .58}41.35 & \cellcolor[rgb]{ 1,  .902,  .608}39.02 & \cellcolor[rgb]{ 1,  .992,  .973}84.40 & \cellcolor[rgb]{ .89,  .953,  .906}46.45 & \cellcolor[rgb]{ 1,  .953,  .812}70.47 & \cellcolor[rgb]{ 1,  .945,  .784}56.87 & \cellcolor[rgb]{ 1,  .973,  .894}86.59 & \cellcolor[rgb]{ 1,  .941,  .773}62.74 & \cellcolor[rgb]{ 1,  .988,  .961}85.56 \\
    WinoGrande & \cellcolor[rgb]{ .675,  .867,  .725}69.93 & \cellcolor[rgb]{ .573,  .824,  .639}73.04 & \cellcolor[rgb]{ .953,  .98,  .961}29.58 & \cellcolor[rgb]{ .545,  .812,  .616}54.58 & \cellcolor[rgb]{ .886,  .953,  .902}86.16 & \cellcolor[rgb]{ .996,  1,  .996}43.54 & \cellcolor[rgb]{ 1,  .984,  .937}76.75 & \cellcolor[rgb]{ .796,  .918,  .827}63.93 & \cellcolor[rgb]{ 1,  .992,  .973}87.94 & \cellcolor[rgb]{ .69,  .871,  .737}75.16 & \cellcolor[rgb]{ .878,  .949,  .898}87.88 \\
    Yelp Polarity & \cellcolor[rgb]{ .8,  .918,  .831}66.97 & \cellcolor[rgb]{ .761,  .902,  .8}65.80 & \cellcolor[rgb]{ 1,  .886,  .541}22.41 & \cellcolor[rgb]{ 1,  .933,  .741}42.42 & \cellcolor[rgb]{ 1,  .973,  .898}80.42 & \cellcolor[rgb]{ 1,  .945,  .788}37.40 & \cellcolor[rgb]{ 1,  .945,  .788}69.25 & \cellcolor[rgb]{ 1,  .953,  .82}57.74 & \cellcolor[rgb]{ .843,  .937,  .867}89.31 & \cellcolor[rgb]{ 1,  .988,  .953}64.98 & \cellcolor[rgb]{ 1,  .973,  .898}82.45 \\
    \bottomrule
    \end{tabular}
    }
    \caption{Target task performances for transferring between intermediate tasks (rows) and target tasks (columns) with RoBERTa as base model. The first row `\textit{No Transfer}' shows the baseline performance when training only on the target task without transfer. All scores are mean values over five random restarts.}
    \label{tab:transfer_results}
    \vspace{-0.5em}
\end{table*}

\section{Methods for the Efficient Selection of Intermediate Tasks}
\label{sec:Methodology}
 
We now present different model selection methods, and later in \S\ref{sec:selection-experiments}, study their effectiveness in our setting outlined above.
We group the different methods based on the 
assumptions they make with regard to the availability of  
intermediate task data $D_S$ and intermediate models $M_S$. 
Access to both can be expensive when considering large pre-trained model repositories with hundreds of tasks.

\subsection{Metadata: Educated Guess}
A setting in which there exist neither access to the intermediate task data $D_S$ nor models trained on the data $M_S$, can be regarded as an \textit{educated guess} scenario. The selection criterion can only rely on metadata available for an intermediate task dataset. 

\vspace{1mm}
\noindent \textbf{Dataset \textit{Size}.} 
Under the  assumption that \textit{more data} implies better transfer performance, the selection criterion denoted as \textit{Size} ranks  
all intermediate tasks in descending order by the training data size.

\vspace{1mm}
\noindent \textbf{Task \textit{Type}.} Under the assumption that similar objective functions 
transfer well, 
we pre-select the subset of tasks of the same type. This approach may be combined with a \textit{random} selection of the remaining tasks, or with ranking them by \textit{size}.

\subsection{ Intermediate Task Data}
With an abundance of available datasets,\footnote{e.g. via \href{https://huggingface.co/datasets}{https://huggingface.co/datasets}.} and the continuous development of new LMs, fine-tuned versions for every task-model combination are not (immediately) available.  
The following methods, thus, leverage the intermediate task data $D_S$ \emph{without} requiring the respective fine-tuned models $M_S$.

\vspace{2mm}
\noindent \textbf{Text Embeddings (\textit{TextEmb}).}
\citet{vuExploringPredictingTransferability2020} pass each example through a  LM and average over the output representations of the final layer (across all examples and all input tokens).  
Assuming that similar embeddings imply positive transfer gains, they rank the intermediate tasks according to their embeddings' cosine similarity to the target task.

\vspace{2mm}
\noindent \textbf{SBERT Embeddings (\textit{SEmb}).} 
Sentence embedding models such as Sentence-BERT \cite[SBERT;][]{reimersSentenceBERTSentenceEmbeddings2019} may be better suited to represent the dataset examples. Similar to \textit{TextEmb}, we rank the intermediate tasks according to their embedding cosine similarity.

\subsection{  Intermediate Model }
Scenarios in which we only have access to the trained intermediate models ($M_S$) occur when the training data is proprietary or if implementing all dataset is too tedious. With the availability of model repositories \cite{wolfTransformersStateoftheArtNatural2020, pfeifferAdapterHubFrameworkAdapting2020} such approaches can be implemented without requiring additional data during model upload (i.e. in contrast to \textit{TaskEmbs}, where the training dataset information needs to be made available). The following describes methods only requiring access to the intermediate models $M_S$. 

\vspace{2mm}
\noindent \textbf{Few-Shot Fine-Tuning (\textit{FSFT}).} 
Fine-tuning of all available intermediate task models on the entire target task is infeasible. As an alternative, we can train models for a few steps on the target task to approximate the final 
performance. 
After $N$ steps on the target task, we rank the intermediate models based on their respective transfer performance.
 
\vspace{2mm}
\noindent \textbf{Proxy Models.}
Following \citet{puigcerverScalableTransferLearning2020}, we leverage simple proxy models to obtain a performance estimation of each trained model $M_S$ on on the \emph{target} dataset $D_T$.
Specifically, we experiment with\textbf{ k-Nearest Neighbors (\textit{kNN})}, with $k=1$ and Euclidian distance, and \textbf{logistic/ linear regression (\textit{linear})} as proxy models.
For both, we first compute $\mathbf{h}_{x_i}^M$, the token-wise averaged output representations of $M_S$, for each training input $x_i \in D_T$.
Using these, we define $D_T^M = \{(\mathbf{h}_{x_i}^M, y_i)\}_{i=1}^N$ as the target dataset embedded by $M_S$.
In the next step, we apply the proxy model on $D_T^M$ and obtain its performance using cross-validation.
By repeating this process for each intermediate task model,
we obtain a list of performance scores
which we leverage to rank the intermediate tasks.


\subsection{Intermediate Model and Task Data }

Access to both intermediate dataset $D_S$ and intermediates model $M_S$ provides a wholesome depiction of the intermediate task, as all previously mentioned methods are applicable in this scenario. 
Further methods which require access to both are: 

\vspace{2mm}
\noindent \textbf{Task Embeddings (\textit{TaskEmb}).} 
\citet{achilleTask2VecTaskEmbedding2019} and \citet{vuExploringPredictingTransferability2020} obtain task embeddings via the Fisher Information Matrix (FIM).  
The FIM captures how sensitive the loss function is towards small perturbations in the weights of the model and thus gives an indication on the importance of certain weights towards solving a task.
 
Given the model weights $\theta$ and the joint distribution of task features and labels $P_\theta(X, Y)$,
we can define the FIM as the expected covariance of the gradients of the log-likelihood w.r.t.  $\theta$:

\vspace{-4mm}
{\small
\begin{align*}
    F_\theta = \mathbb{E}_{x,y \sim P_\theta(X, Y)} \left[ \nabla_\theta \log P_\theta(x, y) \cdot \nabla_\theta \log P_\theta(x, y)^\intercal \right]
\end{align*}
}
We follow the implementation details given in \citet{vuExploringPredictingTransferability2020}. 
For a dataset $D$ and a model $M$ fine-tuned on $D$, we compute the empirical FIM based on $D$'s examples.
The task embeddings are the diagonal entries of the FIM.

\vspace{2mm}
\noindent \textbf{Few-Shot Task Embeddings (\textit{FS-TaskEmb}).}
We also leverage task embeddings in our few-shot scenario outlined above (see \textit{FSFT}), where we fine-tune intermediate models for a few steps on the target dataset. 
With very few training instances, the accuracy scores of FSFT (alone) may not be reliable indicators of the final transfer performances.
As an alternative, we compute the \textit{TaskEmb} similarity of each intermediate model before and after training $N$ steps on the target task.
We then rank all intermediate models in decreasing order of this similarity.

\section{Experimental Setup} 
\label{sec:selection-experiments} 
We evaluate the approaches of \S \ref{sec:Methodology}, each having the objective to rank the intermediate adapters $s \in |\mathcal{S}|$ with respect to their performance on $t \in \mathcal{T}$ when applied in a sequential adapter training setup.
We leverage the transfer performance results of our 462 experiments obtained in \S\ref{sec:transfer} for our ranking task.

\subsection{Hyperparameters}
If not otherwise mentioned, we follow the experimental setup as described in \S\ref{sec:transfer}. We describe method specific hyperparameters in the following.
 
\vspace{1mm}
\noindent \textbf{\textit{SEmb}.} 
We use a Sentence-(Ro)BERT(a)-\textit{base}  
models, fine-tuned on NLI and STS tasks, in concordance with the respective target model type.

\vspace{1mm}
\noindent \textbf{\textit{FSFT}.} 
We fine-tune each  intermediate adapter on the target task for  
one full epoch 
  and rank them based on their target task performances.\footnote{As this represents a rather optimistic estimate of the few-shot transfer performance,  in Appendix \ref{sec:sample_efficiency} we also investigate settings in which we train for
only 5, 10, or 25 update steps.}

\vspace{1mm}
\noindent \textbf{\textit{Proxy Models}.} 
For both \textit{kNN} and \textit{linear}, we obtain performance scores with 5-fold cross-validation 
on each target task.
The architectures slightly vary across task types.
For classification, regression, and multiple-choice target tasks, proxy models predict the label or answer choice.
For sequence tagging tasks, each token in a sequence represents a training instance of $D_T^M$, with the tag being the class label. 
Since this would increase the total number of training examples, we randomly select 1000 embedded examples from $D_T$, to maintain equal sizes of $D_T^M$ across all target tasks.
We do not study proxy models on extractive QA tasks as they cannot directly be transformed into classification tasks.

\vspace{1mm}
\noindent \textbf{\textit{TaskEmb}.} 
We perform standard fine-tuning of randomly initialized adapter modules within the  pre-trained LM to obtain task embeddings. 

\vspace{1mm}
\noindent \textbf{\textit{FS-TaskEmb}.}
We follow the setup of FSFT by training for one epoch (50 update steps). 

\subsection{Metrics}\label{sec:efficient_metrics} 

We compute the \textit{NDCG} \citep{jarvelinCumulatedGainbasedEvaluation2002},
a widely used information retrieval metric that evaluates a ranking with attached relevances (which correspond to our transfer results of \S\ref{sec:transfer}). 

Furthermore, we calculate  \textit{Regret@k} \cite{renggliWhichModelTransfer2020}, which
 measures the relative performance difference between the top $k$ selected intermediate tasks and the optimal intermediate task:
\begin{align*}
\footnotesize
\text{Regret}_k = \frac{
        \overbrace{\max_{s \in \mathcal{S}} \mathbf{E}[T(s, t)]}^{O(\mathcal{S}, t)} - \overbrace{\max_{\hat{s} \in \mathcal{S}_k} \mathbf{E}[T(\hat{s}, t)]}^{M_k(\mathcal{S}, t)}
    }{
        O(\mathcal{S}, t)
    }
\end{align*}
where $T(s, t)$ is the performance on target task $t$ when transferring from intermediate task $s$. 
$O(\mathcal{S}, t)$ denotes the expected target task performance of an optimal selection.
$M_k(\mathcal{S}, t)$ is the highest performance on $t$ among the $k$ top-ranked intermediate tasks of the tested selection method.
We take the difference between both measures and normalize it by the optimal target task performance to obtain our final relative notion of regret.\footnote{We provide more details about our selection of metrics in Appendix \ref{app:metrics}.}

\section{Experimental Results}\label{sec:efficient_results}

\begin{table*}[]
\begin{subtable}[t]{\linewidth}
    \centering
    \def\arraystretch{0.87}
    \resizebox{\linewidth}{!}{
    \begin{tabular}{llrrrrrrrrrrrr|rrr}
\toprule
 & {} & \multicolumn{3}{c}{\textbf{Classification}} & \multicolumn{3}{c}{\textbf{M. Choice}} & \multicolumn{3}{c}{\textbf{QA}} & \multicolumn{3}{c}{\textbf{Tagging}}  & \multicolumn{3}{c}{\textbf{All}} \\
  \cmidrule(lr){3-5} \cmidrule(lr){6-8} \cmidrule(lr){9-11} \cmidrule(lr){12-14} \cmidrule(lr){15-17} 
 & {} &           NDCG &           R@1 &           R@3 &          NDCG &           R@1 &           R@3 &          NDCG &            R@1 &           R@3 &          NDCG &           R@1 &           R@3 &          NDCG &           R@1 &           R@3 \\
\cmidrule(lr){3-17}
\multirow{2}{*}{-} & Random &                      0.49 &                     11.09 &             \textbf{5.55} &             \textbf{0.43} &            \textbf{13.80} &             \textbf{6.30} &                      0.33 &             \textbf{26.95} &            \textbf{18.98} &             \textbf{0.60} &             \textbf{7.65} &             \textbf{2.82} &             \textbf{0.47} &            \textbf{14.09} &             \textbf{7.70} \\
           & Size &             \textbf{0.53} &            \textbf{10.80} &                      6.26 &                      0.39 &                     14.89 &                     11.10 &             \textbf{0.36} &                      33.18 &                     33.18 &                      0.48 &                      8.44 &                      8.44 &                      0.45 &                     15.55 &                     12.87 \\
\cmidrule(lr){3-17}
\multirow{2}{*}{$D_S$} & SEmb &             \textbf{0.82} &             \textbf{0.27} &             \textbf{0.27} &             \textbf{0.79} &                      4.47 & \underline{\textbf{0.00}} & \underline{\textbf{0.49}} &             \textbf{17.04} & \underline{\textbf{7.59}} &                      0.80 &             \textbf{0.47} &                      0.47 &             \textbf{0.75} &             \textbf{4.50} & \underline{\textbf{1.56}} \\
           & TextEmb &                      0.72 &                      2.54 &                      1.20 &                      0.74 &             \textbf{2.94} & \underline{\textbf{0.00}} &                      0.48 &             \textbf{17.04} &                     15.34 &             \textbf{0.88} &             \textbf{0.47} &             \textbf{0.11} &                      0.71 &                      4.91 &                      3.25 \\
\cmidrule(lr){3-17}
\multirow{3}{*}{$M_S$} & FSFT & \underline{\textbf{0.89}} &             \textbf{0.28} & \underline{\textbf{0.00}} & \underline{\textbf{0.89}} & \underline{\textbf{0.00}} & \underline{\textbf{0.00}} &             \textbf{0.28} &             \textbf{21.21} &            \textbf{18.20} & \underline{\textbf{0.97}} & \underline{\textbf{0.00}} & \underline{\textbf{0.00}} & \underline{\textbf{0.79}} &             \textbf{3.96} &             \textbf{3.31} \\
           & kNN &                      0.83 &                      2.49 &                      0.12 &                      0.76 &                      1.91 &                      1.57 &                         - &                          - &                         - &                      0.88 &                      1.44 &                      0.11 &                         - &                         - &                         - \\
           & linear &                      0.79 &                      2.51 &                      1.00 &                      0.89 & \underline{\textbf{0.00}} & \underline{\textbf{0.00}} &                         - &                          - &                         - &                      0.92 &                      0.28 &                      0.28 &                         - &                         - &                         - \\
\cmidrule(lr){3-17}
\multirow{2}{*}{$D_S, M_S$} & FS-TaskEmb &             \textbf{0.87} & \underline{\textbf{0.19}} &             \textbf{0.19} &             \textbf{0.73} &             \textbf{3.03} &             \textbf{0.83} &             \textbf{0.28} & \underline{\textbf{12.90}} &            \textbf{10.38} &             \textbf{0.88} &             \textbf{0.19} &             \textbf{0.19} &             \textbf{0.73} & \underline{\textbf{3.28}} &             \textbf{2.22} \\
           & TaskEmb &                      0.71 &                     14.04 &                      3.08 &                      0.67 &                      6.70 &                      1.92 &                      0.24 &                      30.02 &                     22.40 &                      0.78 &                     31.84 &             \textbf{0.19} &                      0.63 &                     18.18 &                      5.75 \\
\bottomrule
\end{tabular}
    }
    \caption{ RoBERTa}
    \end{subtable}

\vspace{1.0mm}
 
\begin{subtable}[t]{\linewidth}
    \centering
    \def\arraystretch{0.87}
    \resizebox{\linewidth}{!}{
\begin{tabular}{llrrrrrrrrrrrr|rrr}
\toprule
& {} & \multicolumn{3}{c}{\textbf{Classification}} & \multicolumn{3}{c}{\textbf{M. Choice}} & \multicolumn{3}{c}{\textbf{QA}} & \multicolumn{3}{c}{\textbf{Tagging}}  & \multicolumn{3}{c}{\textbf{All}} \\
  \cmidrule(lr){3-5} \cmidrule(lr){6-8} \cmidrule(lr){9-11} \cmidrule(lr){12-14} \cmidrule(lr){15-17} 
 & {} &           NDCG &           R@1 &           R@3 &          NDCG &           R@1 &           R@3 &          NDCG &            R@1 &           R@3 &          NDCG &           R@1 &           R@3 &          NDCG &           R@1 &           R@3 \\
\cmidrule(lr){3-17}
\multirow{2}{*}{-} & Random &                      0.45 &             \textbf{8.61} &                      6.04 &             \textbf{0.50} &             \textbf{8.40} &             \textbf{5.03} &             \textbf{0.44} &            \textbf{29.35} &            \textbf{18.65} &             \textbf{0.56} &             \textbf{7.26} &             \textbf{2.95} &                      0.48 &            \textbf{12.08} &             \textbf{7.50} \\
           & Size &             \textbf{0.51} &                     11.34 &             \textbf{5.87} &                      0.50 &                     11.85 &                      7.51 &                      0.42 &                     33.80 &                     33.80 &                      0.48 &                      9.37 &                      9.37 &             \textbf{0.48} &                     15.20 &                     12.03 \\
\cmidrule(lr){3-17}
\multirow{2}{*}{$D_S$} & SEmb &                      0.78 &             \textbf{0.75} &             \textbf{0.53} &                      0.59 &                      7.93 &             \textbf{1.25} & \underline{\textbf{0.88}} & \underline{\textbf{2.98}} & \underline{\textbf{0.00}} &             \textbf{0.79} &             \textbf{0.56} &                      0.56 &                      0.75 &                      3.08 & \underline{\textbf{0.63}} \\
           & TextEmb &             \textbf{0.81} &                      1.26 &                      0.75 &             \textbf{0.60} &             \textbf{6.77} &                      1.46 &                      0.86 & \underline{\textbf{2.98}} &                      2.42 &                      0.79 &             \textbf{0.56} &             \textbf{0.51} &             \textbf{0.76} & \underline{\textbf{2.95}} &                      1.20 \\
\cmidrule(lr){3-17}
\multirow{3}{*}{$M_S$} & FSFT & \underline{\textbf{0.93}} & \underline{\textbf{0.33}} &                      0.33 &                      0.72 &                      4.16 &                      1.64 &             \textbf{0.39} &            \textbf{17.07} &            \textbf{17.07} &                      0.89 &                      0.65 &                      0.50 &             \textbf{0.77} &             \textbf{4.48} &             \textbf{3.76} \\
           & kNN &                      0.90 &                      1.10 & \underline{\textbf{0.00}} &                      0.68 &                      2.82 &                      1.85 &                         - &                         - &                         - &                      0.94 & \underline{\textbf{0.00}} & \underline{\textbf{0.00}} &                         - &                         - &                         - \\
           & linear &                      0.82 &                      3.66 &                      1.86 & \underline{\textbf{0.76}} & \underline{\textbf{1.35}} & \underline{\textbf{0.86}} &                         - &                         - &                         - & \underline{\textbf{0.96}} & \underline{\textbf{0.00}} & \underline{\textbf{0.00}} &                         - &                         - &                         - \\
\cmidrule(lr){3-17}
\multirow{2}{*}{$D_S, M_S$} & FS-TaskEmb &             \textbf{0.92} &             \textbf{0.62} & \underline{\textbf{0.00}} &             \textbf{0.72} &                      5.38 &             \textbf{0.93} &                      0.66 &                     11.17 &             \textbf{2.07} &             \textbf{0.82} &                      1.37 &             \textbf{0.50} & \underline{\textbf{0.80}} &                      3.97 &             \textbf{0.72} \\
           & TaskEmb &                      0.83 &                      3.89 &                      2.02 &                      0.72 &             \textbf{4.19} &                      1.17 &             \textbf{0.67} &             \textbf{3.61} &                      3.61 &                      0.73 &             \textbf{1.36} &             \textbf{0.50} &                      0.75 &             \textbf{3.46} &                      1.80 \\
\bottomrule
\end{tabular}
    }
    \caption{ BERT}
    \end{subtable}
    \vspace{-0.8em}
        \caption{Evaluation of intermediate task rankings produced by different methods for RoBERTa (a) and BERT (b).
    The table shows the mean \textit{NDCG} and \textit{Regret} scores by target task type.
    The best score in each group is highlighted in bold, best overall score is underlined.
    For \textit{NDCG}, higher is better; for \textit{Regret}, lower is better.}
    \label{tab:method_results}
    \vspace{-1em}
\end{table*}

Table~\ref{tab:method_results} shows the results when selecting among all available intermediate tasks for BERT and RoBERTa.\footnote{ Table~\ref{tab:method_results_task_type} in the appendix shows results when preferring tasks of the same \textit{type} for BERT and RoBERTa.} As expected the \textit{Random} and \textit{Size} baselines do not yield good rankings when selecting among all intermediate tasks.

\vspace{1mm}
\noindent\textbf{Access to only $\mathbf{D}_{\mathbf{S}}$ or $\mathbf{M}_{\mathbf{S}}$.}
These methods typically perform better than our baselines.
 
TextEmb and SEmb perform on par in most cases.\footnote{The used SBERT model is trained on NLI and STS-B tasks, which are included in our set of intermediate and target tasks, respectively. A direct comparison between TextEmb and SEmb for the respective classification tasks is thus difficult. } 
While FSFT outperforms the other approaches in most cases, it comes at the high cost of requiring downloading and fine-tuning all intermediate models for a few steps. This can be prohibitive if we consider many intermediate tasks. If we have access to TextEmb or SEmb information of the intermediate task (i.e., individual vectors  distributed as part of a model repository), these techniques yield similar performances at a much lower cost.

\vspace{1mm}
\noindent\textbf{Access to both $\mathbf{D}_{\mathbf{S}}$ and $\mathbf{M}_{\mathbf{S}}$.}
Assuming the availability of \emph{both} intermediate models and intermediate data is the most prohibitive setting.
Surprisingly, we find BERT and RoBERTa to behave considerably differently, especially evident for QA tasks. As shown by \citet{vuExploringPredictingTransferability2020}, TaskEmb performs very well for BERT, however we find that the results of this gradient based approach do not translate to RoBERTa. While these approaches perform best or competitively for all task types using BERT, they considerably underperform all methods when leveraging pre-trained RoBERTa weights. Here, the two much simpler domain embedding methods outperform the \textit{TaskEmb} method based on the FIM.

\vspace{1mm}
\noindent\textbf{Summary.} We find that simple indicators such as domain similarity are suitable for selecting intermediate pre-training tasks for both BERT and RoBERTa based models. 
Our evaluated methods are able to efficiently select the best performing intermediate tasks with a \textit{Regret@3} of 0.0 in many cases.  
Our results, thus, show that the selection methods are able to effectively rank the top tasks with relative certainty, thus considerably reducing the number of necessary experiments.\footnote{We also find that combining domain and task type match indicators often yield the best overall results, outperforming computationally more expensive methods. See Appendix~\ref{app:task_type} for more experiments with task type pre-selection. }

\section{Analysis}\label{sec:efficient_analysis}

\noindent\textbf{Computational Costs.}
Table~\ref{table:complexity} estimates the computational costs of 
each 
transfer source selection method. 
\emph{Complexity} shows the required data passes through the model.\footnote{We neglect computations related to embedding similarities and proxy models as they are cheap compared to model forward/ backward passes.}
For the embedding-based approaches, we assume pre-computed embeddings for all intermediate tasks. 
For TaskEmb, we only train an adapter on the target task for $e$ epochs. 

In addition to the complexity, we calculate the required Multiply-Accumulate computations (MAC) for 42 intermediate tasks and one target task with 1000 training examples, each with an average sequence length of 128.\footnote{ We recorded MAC with the \href{https://github.com/Lyken17/pytorch-OpCounter/}{pytorch-OpCounter} package. 
}
Following our experimental setup in §\ref{sec:selection-experiments}, we set $e = 15$ for TaskEmb and $e = 1$ for FSFT/ FS-TaskEmb.
We find that embedding-based methods require two orders of magnitude fewer computations compared to fine-tuning approaches.
The difference may be even larger when we consider more intermediate tasks. Since fine-tuning approaches do not yield gains that would warrant the high computational expense (see \S\ref{sec:efficient_results}), we conclude that \textit{SEmb} has the most favorable trade-off between  efficiency and effectiveness.

\begin{table}[t]
    \centering
    \def\arraystretch{0.87}
    \resizebox{\linewidth}{!}{
    \begin{tabular}{lcr}
    \toprule
    \bf Method & \bf Complexity & \bf MACs \\
    \midrule
    Metadata & $1$  & 0 \\
    TextEmb/ SEmb & $f$  & 1.10E+13 \\
    TaskEmb & $(e + 1) f + e b$ & 3.30E+14 \\
    kNN/ linear & $n f$ & 4.61E+14 \\
    FSFT/ FS-TaskEmb & $2 n e f + n e b$ & 1.38E+15 \\
    \bottomrule
    \end{tabular}%
    }
    \caption{Computational cost of transfer source selection.
    $f$ denotes a forward pass through all target task examples once, $b$ is the corresponding backward pass, $n$ is the number of source models, and $e$ is the number of full training epochs (for FS approaches $e\leq 1$).
    }
    \label{table:complexity}
    \vspace{-0.5em}
\end{table}

\vspace{1mm}
\noindent\textbf{SEmb Model Dependency.}
We compare different pre-trained sentence-embedding model variants to identify the extent to which SEmb is invariant to such changes. 
We experiment with BERT and RoBERTa variants of sizes \textit{Distill}, \textit{Base}, and \textit{Large}, and present results for RoBERTa tasks in Table~\ref{tab:SEmb_types_all}.\footnote{The full results can be found in Table~\ref{tab:SEmb_types} of the appendix.} We find that all variants perform comparably, demonstrating that SEmb is a computationally efficient, model-type invariant method for selecting beneficial intermediate tasks.

\begin{table}[]
    \centering
    \footnotesize
    \def\arraystretch{0.95}
    \begin{tabularx}{0.97\linewidth}{l X X X X}
\toprule
{} &           NDCG &           R@1 &           R@3 &       R@5 \\
\midrule
SEmb-BERT$_D$    &          0.72 &          5.50 &          2.07 & 1.69\\
SEmb-BERT$_B$    &          0.72 &          4.99 &          1.16 & \textbf{0.07}\\
SEmb-BERT$_L$    &          0.70 &          6.30 &          2.12 & 1.01\\
SEmb-RoBERTa$_D$ & \textbf{0.77} &          4.60 &          0.82 & 0.44\\
SEmb-RoBERTa$_B$ &          0.75 &          4.50 &          1.56 & 0.48\\
SEmb-RoBERTa$_L$ &          0.74 & \textbf{3.96} & \textbf{0.47} & \textbf{0.07}\\
\bottomrule
\end{tabularx}
    
    \caption{Intermediate SEmb rankings for RoBERTa tasks produced by different model-type variants. \textit{D, B,} and \textit{L} stand for \textit{Distill}, \textit{Base}, and \textit{Large}, respectively.
    The table shows the mean \textit{NDCG} and \textit{Regret} scores.
    For \textit{NDCG}, higher is better; for \textit{Regret}, lower is better.}
    \label{tab:SEmb_types_all}
    \vspace{-1em}
\end{table}

\begin{figure}
    \centering
    \includegraphics[width=\linewidth,trim=0cm 0.5cm 0cm 0.5cm,clip]{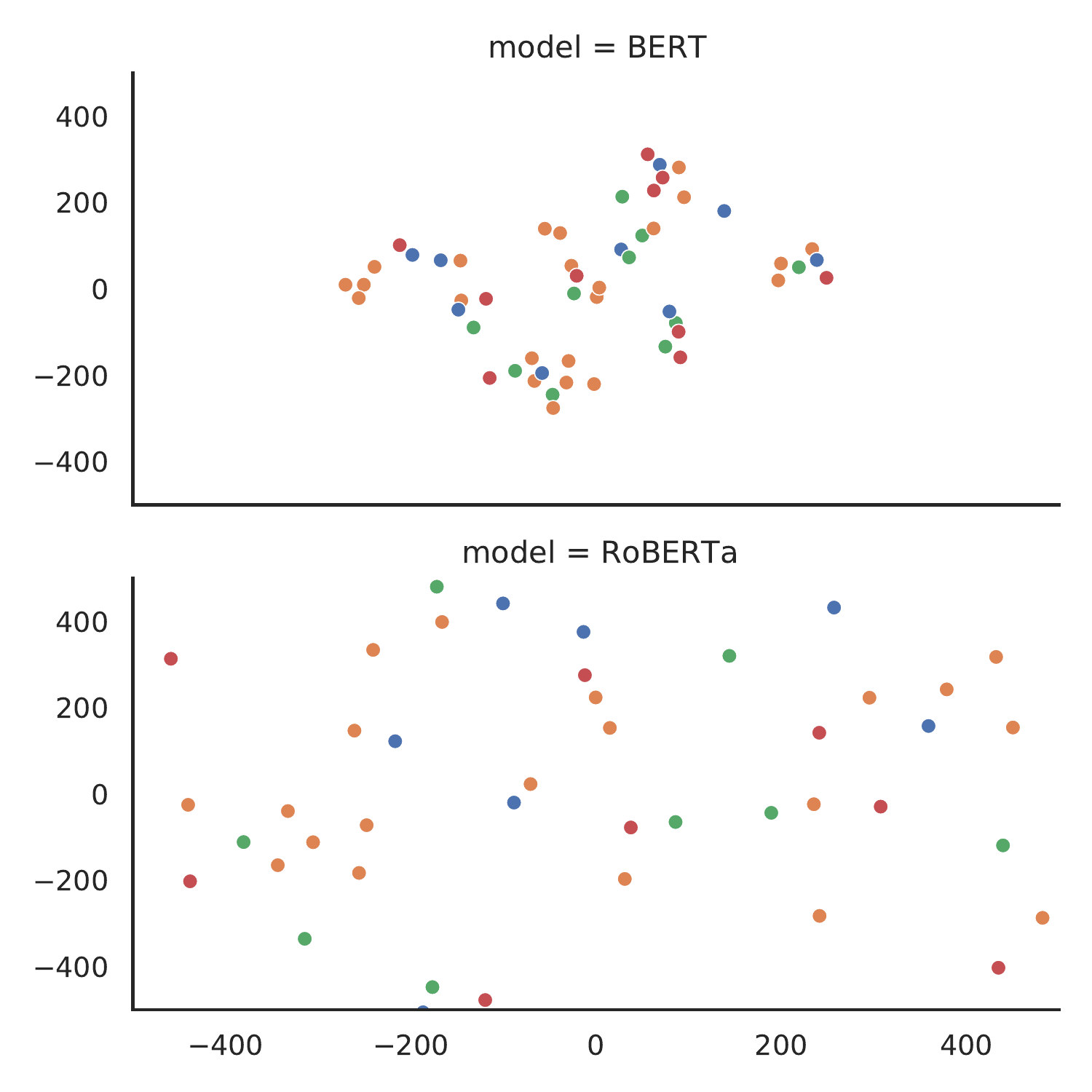}
    \caption{Clustering of  BERT and RoBERTa TaskEmbs, respectively, using T-SNE. Colors indicate task types. We compared different random seeds, all of which resulted in similar visualizations.}
    \label{fig:tsne}
\vspace{-1.5mm}
\end{figure}

\vspace{1mm}
\noindent\textbf{BERT vs RoBERTa TaskEmb Space.}
To better understand the TaskEmb performance differences between BERT and RoBERTa models, we visualize the respective embedding spaces using T-SNE in Figure~\ref{fig:tsne}. 
We find that BERT embeddings are clustered much more closely in the vector space than RoBERTa embeddings. While TaskEmbs of BERT also seem to be located in the proximity of related tasks, 
TaskEmbs of RoBERTa are distributed further apart. This can result in worse performance due to the curse of dimensionality. 

Overall, our results and analysis suggest that TaskEmb, unlike SentEmb, considerably depend on the chosen base model.

\begin{figure}
    \centering
    \includegraphics[width=\linewidth]{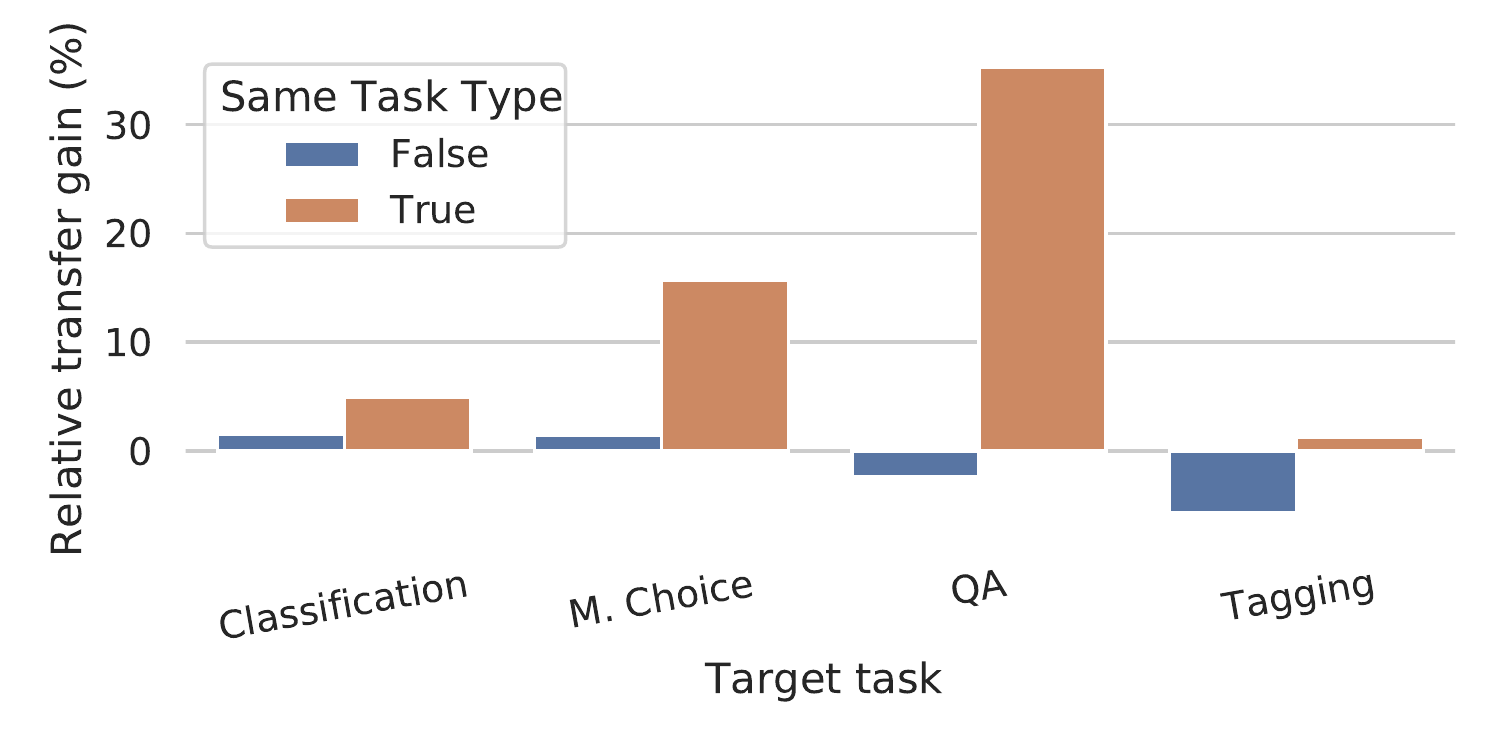}
    \caption{Relative transfer gains for transfer within and across types, split by target task type. Results shown for RoBERTa.}
    \label{fig:transfer_task_type}
\end{figure}

\vspace{1mm}
\noindent\textbf{Within- and Across-Type Transfer.}
Our experimental setup includes tasks of four different types, i.e. Transformer prediction head structures: \textit{sequence classification/ regression}, \textit{multiple choice}, \textit{extractive question answering} and \textit{sequence tagging}.
Figure~\ref{fig:transfer_task_type} compares the relative transfer gains within and across these task types for RoBERTa.
We see that within-type transfer is consistently stronger across all target tasks. 
We find the largest differences between within-type and across-type transfer 
for the extractive QA target tasks.
These observations may be partly explained by the homogeneity of the included QA intermediate tasks;
They overwhelmingly focus on general reading comprehension across multiple domains with paragraphs from Wikipedia or the web as contexts.
Tasks of other types more distinctly focus on individual domains and scenarios.

Overall, we find a negative across-type transfer gain (i.e., loss) for \emph{8 out of 11 tested target tasks} (on average).
This suggests that task type match between intermediate and target task is a strong indicator for transfer success.
Thus, in the next section, we evaluate variants of all methods presented in §\ref{sec:Methodology} that prefer intermediate tasks of the same type as the target task. 

\vspace{1mm}
\noindent\textbf{Pre-Ranking by Task Types.}
We implement a simple mechanism to ensure that tasks with the same type as the target task are always ranked before tasks of other types during intermediate task selection.
Given a task selection method, we first rank all tasks of the same type at the top before ranking tasks of all other types below.
Results for applying this mechanism to all presented task selection methods are given for BERT and RoBERTa in Table~\ref{tab:method_results_task_type} of the Appendix.

We find that even though the random and \textit{Size} baselines do not yield good rankings when selecting among all intermediate tasks (cf. Table~\ref{tab:method_results}), the scores considerably improve when preferring tasks of the same type.
In general, we see almost consistent improvements across all task selection methods for both BERT and RoBERTa when implementing pre-ranking by task types.
Considering all target tasks and all methods, preferring intermediate tasks of the same type yields improved NDCG scores in 77 of 99 cases.  

\vspace{1mm}
\noindent\textbf{Further Analysis.}  
We further find that embedding based approaches are sample efficient, while FSFT appproaches are not (\S \ref{sec:sample_efficiency}). We also report results for combining ranking approaches with Rank Fusion, which does not yield consistent improvements over the individual approaches presented before (\S \ref{subsec:efficient_method_combinations}).

\section{Conclusion}

In this work 
we have established that intermediate pre-training \textit{can} yield gains in adapter-based setups, 
however, 
around 44\% of all transfer combinations result in decreased performances. 
We have consolidated several existing and new methods for efficiently identifying beneficial intermediate tasks.
Experimenting with different model types, we find that the previously proposed best performing approaches for BERT do not translate to RoBERTa. 

Overall, efficient embedding based methods, such as those relying on pre-computable sentence representations, perform better or often on-par with more expensive approaches. The best methods achieve a \textit{Regret@3} of less than 1\% on average, demonstrating that they are effective at efficiently identifying the best intermediate tasks. The approaches evaluated and proposed in this work, thus, enable the automatic identification of beneficial intermediate tasks, deeming exhaustive experimentation on many task-combinations unnecessary. 
When applied on a broad scale, these methods can contribute to more sustainable \cite{strubell-etal-2019-energy,moosavi2020proceedings} and more inclusive \cite{joshi-etal-2020-state} natural language processing. 

\section*{Acknowledgements}
Clifton and Jonas  are supported by the LOEWE initiative (Hesse, Germany) within the emergenCITY center.  
Andreas was supported by the German Research Foundation (DFG) as part of the UKP-SQuARE project (grant GU 798/29-1).

We thank Leonardo Ribeiro and the anonymous reviewers for insightful feedback and suggestions on a draft of this paper.
 
\bibliography{anthology,thesis}
\bibliographystyle{acl_natbib}

\appendix

\section{Tasks}\label{app:tasks}

Our experiments cover a diverse set of 53 different tasks, broadly divided into the four task types \textit{sequence classification/ regression}, \textit{multiple choice}, \textit{extractive question answering} and \textit{sequence tagging}.
Motivated by previous work, we first select tasks that are either part of widely used benchmarks \citep{wangGLUEMultiTaskBenchmark2018, wangSuperGLUEStickierBenchmark2019, talmorMultiQAEmpiricalInvestigation2019} or have been successfully applied to sequential transfer setups previously \citep{sapSocialIQACommonsense2019, liuLinguisticKnowledgeTransferability2019, pruksachatkunIntermediateTaskTransferLearning2020, vuExploringPredictingTransferability2020}.
Additionally, we include other recent challenging tasks that fall under the four defined task types (e.g. \citet{bhagavatulaAbductiveCommonsenseReasoning2019,rogersGettingCloserAI2020}) and tasks that extend the range of included dataset sizes and task domains.
In general, we focus on tasks with publicly available datasets, e.g. via \textit{HuggingFace Datasets}\footnote{\url{https://huggingface.co/datasets}}.
Our full set of tasks is split into 42 intermediate tasks, presented in Table~\ref{table:source_tasks}, and 11 target tasks, presented in Table~\ref{table:target_tasks}.

\section{Transfer training details}\label{app:training}

For all our experiments, we use the PyTorch implementations of BERT and RoBERTa in the \textit{HuggingFace Transformers} library \citep{wolfTransformersStateoftheArtNatural2020} as the basis.
The adapter implementation is provided by the \textit{AdapterHub} framework \citep{pfeifferAdapterHubFrameworkAdapting2020} and integrated into the Transformers library \footnote{\url{https://github.com/Adapter-Hub/adapter-transformers}}.

In the light of the number and variety of different tasks used, we don't perform any extensive hyperparameter tuning on each training task.
We mostly adhere to the hyperparameter recommendations of the Transformers library and \citet{pfeifferAdapterFusionNonDestructiveTask2020} for adapter training.
Specifically, we train all adapters for a maximum of 15 epochs, with early stopping after 3 epochs without improvements on the validation set.
We use a learning rate of $10^{-4}$ and batch sizes between $4$ and $32$, depending on the size of the dataset.
These settings apply to the adapter training on each intermediate task as well as the subsequent fine-tuning on the target dataset.
Additionally, since performances on the low-resource target tasks can be unstable, we perform multiple random restarts (five restarts for RoBERTa and three restarts for BERT) for all training runs on the target tasks, reporting the mean of all restarts.
The final scores on each task are computed on the respective tests set if publicly available, otherwise on the validation sets.

Results for RoBERTa are shown in Table~\ref{tab:transfer_results} and results for BERT are shown in Table~\ref{tab:transfer_results_BERT}.

\section{Metrics for transfer source selection}\label{app:metrics}

\subsection{NDCG}

Following \citet{vuExploringPredictingTransferability2020}, we compute the \textit{Normalized Discounted Cumulative Gain (NDCG)} \citep{jarvelinCumulatedGainbasedEvaluation2002},
a widely used information retrieval metric that evaluates a ranking with attached relevances.
The NDCG is defined via the \textit{Discounted Cumulative Gain (DCG)}, which represents a relevance score for a set of items, each discounted by its position in the ranking.
The DCG of a ranking $R$, accumulated at a particular rank position $p$, can be computed as:

\begin{align*}
    \text{DCG}_p(R) = \sum_{i=1}^p \frac{2^{\text{rel}_i} - 1}{\log_2(i+1)}
\end{align*}

In our setting, $R$ refers to a ranking of intermediate tasks where the relevance $\text{rel}_i$ of the intermediate task with rank $i$ is set to the mean target performance when transferring the adapter trained on this intermediate task, i.e. $\text{rel}_i \in [0, 100]$.
We always evaluate the full ranking of intermediate tasks, thus we set $p = |\mathcal{S}|$.

The NDCG finally normalizes the DCG of the ranking predicted by the task selection method ($R_{pred}$) by the perfect ranking produced by the empirical transfer results ($R_{true}$).
An NDCG of $100\%$ indicates a perfect ranking.

\begin{align*}
    \text{NDCG}_p(R) = \frac{\text{DCG}_p(R_{pred})}{\text{DCG}_p(R_{true})}
\end{align*}

\subsection{Choice of metrics}

Our selection of evaluation metrics combines two measures that both evaluate the quality of the full ranking (\textit{NDCG}) and the top selections of each methods (\textit{Regret}).
We prefer this combination of metrics over various other common possible evaluation metrics.
We experimented with classical correlation measures such as Spearman rank correlation, finding they give poor indication on the overall quality of a selection method.
The Spearman correlation is agnostic to the location within the ranking, thus penalizing mismatches at the bottom of the ranking with the same weight as mismatches at the top.
In our setting, the top ranks are more important, making the NDCG which is biased towards correct rankings at the top a better fit.
\citet{renggliWhichModelTransfer2020} further discuss the limitations of correlation as an evaluation metric for task selection.

\citet{vuExploringPredictingTransferability2020} use the average predicted rank $\rho$ of the source task with the best target performance as an additional metric.
However, this metric does not account for the real target performance difference between the top ranked source tasks across different methods.
In a simple example, assume two selection methods $A$ and $B$ assign the top performing source task $s_{max}$ to the same average rank.
Further, $A$ ranks a different source task on top which nearly performs on par with $s_{max}$ while $B$ predicts a much weaker source task on top.
In this case, we clearly would want to prefer method $A$ over method $B$.
Unlike $\rho$, our choice of regret as evaluation metric considers these differences.

\begin{figure}[]
    \centering
    \includegraphics[width=\linewidth]{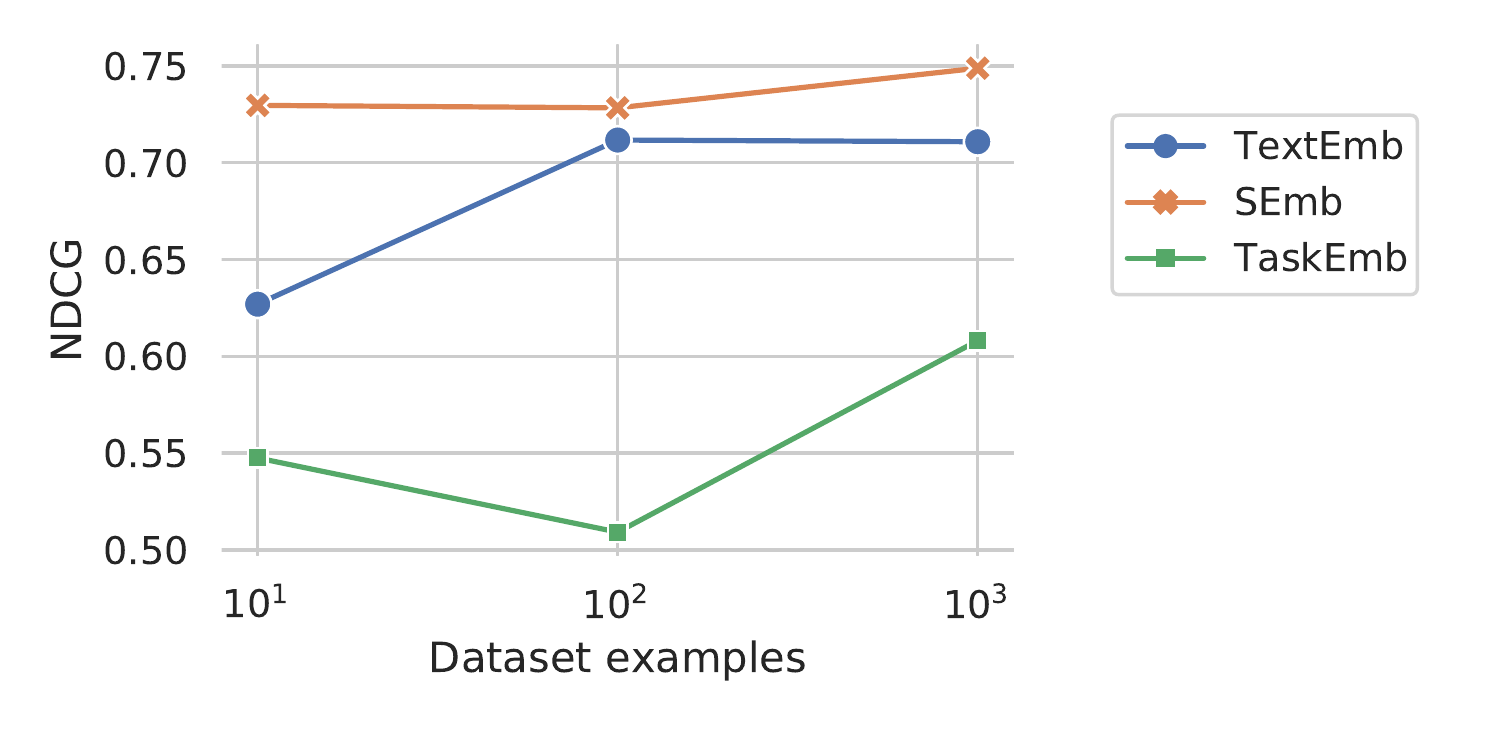}
    \caption{Intermediate task selection performances for feature embedding methods with different data sizes on the target task. Results shown for RoBERTa and averaged over all targets.}
    \label{fig:ranking_sizes}
\end{figure}

\section{Sample Efficiency}
\label{sec:sample_efficiency}
\noindent \textbf{Embedding-based approaches.} 
Intermediate pre-training can have a larger impact on small target tasks. We therefore analyze and compare the effectiveness of embedding-based approaches with  
only $10$, $100$, and $1000$ target examples. 
 
Figure~\ref{fig:ranking_sizes} plots the results for all feature embedding methods when applied to intermediate task selection for RoBERTa.
We find that the quality of the rankings can decrease substantially in the smallest setting with only 10 target examples. \emph{SEmb} is a notable exception, achieving results close to that of the full 1000 examples ($73\%$ vs. $74.9\%$ NDCG). With that, SEmb consistently performs above all other methods in all settings. 

\begin{figure}[]
    \centering
    \includegraphics[width=\linewidth]{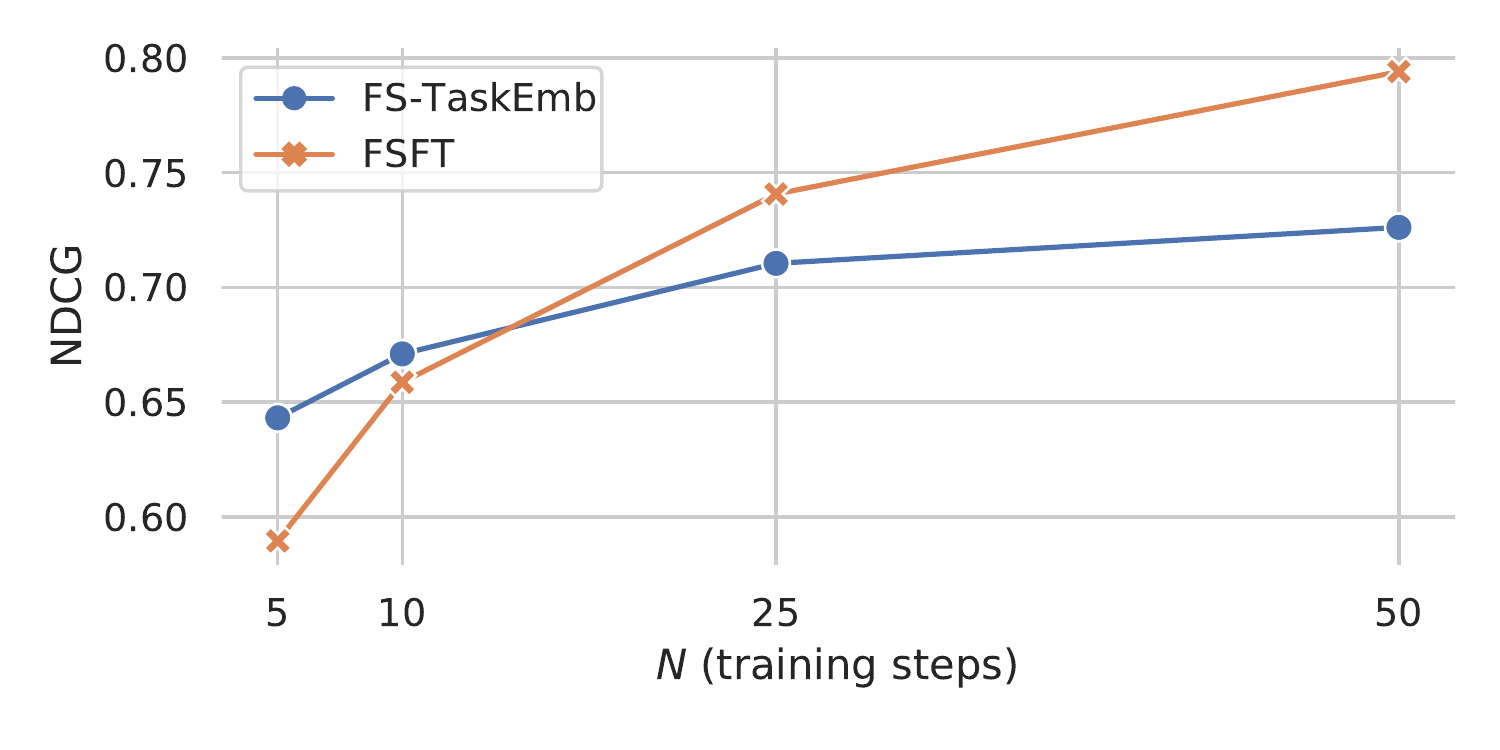}
    \caption{Intermediate task selection performances for fine-tuning methods at different checkpoints. Results shown for RoBERTa and averaged over all targets.}
    \label{fig:ranking_probe}
\end{figure}

\vspace{1mm}
\noindent \textbf{Few-Shot approaches.} 
We experiment with  $\text{N} \in \{5, 10, 25, 50 \}$ update steps for the fine-tuning methods \textit{FSFT} and \textit{FS-TaskEmb}.
Results for RoBERTa are shown in Figure~\ref{fig:ranking_probe}.
While unsurprisingly, the performance for both methods improves consistently  with the number of fine-tuning steps, \textit{FS-TaskEmb} produces superior rankings at earlier checkpoints, however is outperformed by \textit{FSFT} on the long run.
The results indicate that updating  for $<25$ update steps does not provide sufficient  evidence to reliably predict the best intermediate tasks.

\begin{table*}[]
\begin{subtable}[t]{\linewidth}
    \centering
    \def\arraystretch{0.87}
    \resizebox{\linewidth}{!}{
    \begin{tabular}{llrrrrrrrrrrrr|rrr}
\toprule
 & {} & \multicolumn{3}{c}{\textbf{Classification}} & \multicolumn{3}{c}{\textbf{M. Choice}} & \multicolumn{3}{c}{\textbf{QA}} & \multicolumn{3}{c}{\textbf{Tagging}}  & \multicolumn{3}{c}{\textbf{All}} \\
  \cmidrule(lr){3-5} \cmidrule(lr){6-8} \cmidrule(lr){9-11} \cmidrule(lr){12-14} \cmidrule(lr){15-17} 
 & {} &           NDCG &           R@1 &           R@3 &          NDCG &           R@1 &           R@3 &          NDCG &            R@1 &           R@3 &          NDCG &           R@1 &           R@3 &          NDCG &           R@1 &           R@3 \\
\cmidrule(lr){3-17}
\multirow{2}{*}{-} & Random-T &             \textbf{0.54} &             \textbf{8.81} &             \textbf{5.09} &                      0.66 &                      5.50 &                      1.81 &                      0.57 &                      9.46 &                      3.78 &                      0.84 &                      1.54 &             \textbf{0.38} &                      0.63 &                      6.71 &                      3.10 \\
           & Size-T &                      0.54 &                     10.80 &                      6.26 &             \textbf{0.66} &             \textbf{4.30} &             \textbf{1.22} & \underline{\textbf{0.96}} & \underline{\textbf{0.00}} & \underline{\textbf{0.00}} &             \textbf{0.87} &             \textbf{0.47} &                      0.47 &             \textbf{0.71} &             \textbf{5.18} &             \textbf{2.69} \\
\cmidrule(lr){3-17}
\multirow{2}{*}{$D_S$} & SEmb-T &             \textbf{0.83} &             \textbf{0.27} &                      0.27 &             \textbf{0.92} & \underline{\textbf{0.00}} & \underline{\textbf{0.00}} &                      0.54 &                      7.76 &                      4.04 &                      0.94 &             \textbf{0.47} & \underline{\textbf{0.00}} &             \textbf{0.82} & \underline{\textbf{1.59}} &                      0.83 \\
           & TextEmb-T &                      0.75 &                      2.13 &             \textbf{0.19} &                      0.89 &                      0.38 & \underline{\textbf{0.00}} &             \textbf{0.62} &             \textbf{4.04} &             \textbf{2.05} &             \textbf{0.95} &             \textbf{0.47} & \underline{\textbf{0.00}} &                      0.80 &                      1.70 & \underline{\textbf{0.44}} \\
\cmidrule(lr){3-17}
\multirow{3}{*}{$M_S$} & FSFT-T &             \textbf{0.86} &             \textbf{0.28} & \underline{\textbf{0.00}} &                      0.93 & \underline{\textbf{0.00}} & \underline{\textbf{0.00}} &             \textbf{0.49} &            \textbf{10.99} &            \textbf{10.39} & \underline{\textbf{0.97}} & \underline{\textbf{0.00}} & \underline{\textbf{0.00}} & \underline{\textbf{0.83}} &             \textbf{2.10} &             \textbf{1.89} \\
           & kNN-T &                      0.82 &                      2.49 &                      0.12 &                      0.81 &                      1.91 &                      1.91 &                         - &                         - &                         - &                      0.95 &                      0.11 &                      0.11 &                         - &                         - &                         - \\
           & linear-T &                      0.78 &                      1.84 &                      1.49 & \underline{\textbf{0.96}} & \underline{\textbf{0.00}} & \underline{\textbf{0.00}} &                         - &                         - &                         - &                      0.95 & \underline{\textbf{0.00}} & \underline{\textbf{0.00}} &                         - &                         - &                         - \\
\cmidrule(lr){3-17}
\multirow{2}{*}{$D_S, M_S$} & FS-TaskEmb-T & \underline{\textbf{0.88}} & \underline{\textbf{0.19}} & \underline{\textbf{0.00}} &                      0.75 &             \textbf{3.03} &                      0.83 &             \textbf{0.46} &            \textbf{12.90} &             \textbf{4.19} &             \textbf{0.93} &             \textbf{0.19} &             \textbf{0.19} &             \textbf{0.78} &             \textbf{3.28} &             \textbf{1.02} \\
           & TaskEmb-T &                      0.76 &                      4.82 &                      0.12 &             \textbf{0.76} &                      3.74 &             \textbf{0.60} &                      0.45 &            \textbf{12.90} &                      5.42 &                      0.92 &             \textbf{0.19} &             \textbf{0.19} &                      0.73 &                      5.15 &                      1.23 \\
\bottomrule
\end{tabular}
    }
    \caption{ RoBERTa}
    \end{subtable}

\vspace{1.0mm}
 
\begin{subtable}[t]{\linewidth}
    \centering
    \def\arraystretch{0.87}
    \resizebox{\linewidth}{!}{
\begin{tabular}{llrrrrrrrrrrrr|rrr}
\toprule
& {} & \multicolumn{3}{c}{\textbf{Classification}} & \multicolumn{3}{c}{\textbf{M. Choice}} & \multicolumn{3}{c}{\textbf{QA}} & \multicolumn{3}{c}{\textbf{Tagging}}  & \multicolumn{3}{c}{\textbf{All}} \\
  \cmidrule(lr){3-5} \cmidrule(lr){6-8} \cmidrule(lr){9-11} \cmidrule(lr){12-14} \cmidrule(lr){15-17} 
 & {} &           NDCG &           R@1 &           R@3 &          NDCG &           R@1 &           R@3 &          NDCG &            R@1 &           R@3 &          NDCG &           R@1 &           R@3 &          NDCG &           R@1 &           R@3 \\
\cmidrule(lr){3-17}
\multirow{2}{*}{-} & Random-T &                      0.50 &             \textbf{8.88} &             \textbf{5.58} &                      0.58 &                      6.14 &                      3.31 &                      0.72 &                      8.49 &                      2.95 &                      0.74 &                      2.02 &                      0.91 &                      0.61 &                      6.82 &                      3.63 \\
           & Size-T &             \textbf{0.53} &                     11.34 &                      5.87 &             \textbf{0.60} &             \textbf{5.97} &             \textbf{1.78} &             \textbf{0.84} &             \textbf{3.61} & \underline{\textbf{0.00}} &             \textbf{0.74} &             \textbf{1.36} &             \textbf{0.85} &             \textbf{0.64} &             \textbf{6.65} &             \textbf{2.78} \\
\cmidrule(lr){3-17}
\multirow{2}{*}{$D_S$} & SEmb-T &                      0.82 &             \textbf{0.75} &             \textbf{0.31} & \underline{\textbf{0.74}} & \underline{\textbf{0.93}} & \underline{\textbf{0.93}} & \underline{\textbf{0.89}} &             \textbf{2.98} & \underline{\textbf{0.00}} &             \textbf{0.83} & \underline{\textbf{0.56}} &             \textbf{0.51} &             \textbf{0.81} & \underline{\textbf{1.17}} & \underline{\textbf{0.46}} \\
           & TextEmb-T &             \textbf{0.82} &                      1.26 &                      0.75 &                      0.72 &                      1.95 & \underline{\textbf{0.93}} &                      0.87 &             \textbf{2.98} &                      2.42 &                      0.82 & \underline{\textbf{0.56}} &             \textbf{0.51} &                      0.80 &                      1.63 &                      1.06 \\
\cmidrule(lr){3-17}
\multirow{3}{*}{$M_S$} & FSFT-T &             \textbf{0.92} &             \textbf{0.33} & \underline{\textbf{0.00}} &                      0.73 &                      5.38 &                      1.95 &             \textbf{0.78} & \underline{\textbf{0.00}} & \underline{\textbf{0.00}} &                      0.88 &                      0.65 & \underline{\textbf{0.50}} & \underline{\textbf{0.84}} &             \textbf{1.70} &             \textbf{0.62} \\
           & kNN-T &                      0.91 &                      1.10 & \underline{\textbf{0.00}} &                      0.70 &             \textbf{2.82} &                      1.46 &                         - &                         - &                         - & \underline{\textbf{0.88}} & \underline{\textbf{0.56}} &                      0.51 &                         - &                         - &                         - \\
           & linear-T &                      0.79 &                      3.00 &                      1.70 &             \textbf{0.73} &                      2.94 & \underline{\textbf{0.93}} &                         - &                         - &                         - &                      0.85 &                      0.91 &                      0.51 &                         - &                         - &                         - \\
\cmidrule(lr){3-17}
\multirow{2}{*}{$D_S, M_S$} & FS-TaskEmb-T & \underline{\textbf{0.95}} & \underline{\textbf{0.00}} & \underline{\textbf{0.00}} &                      0.71 &                      5.38 & \underline{\textbf{0.93}} &                      0.67 &                     11.17 &                      2.07 &             \textbf{0.80} &                      1.37 & \underline{\textbf{0.50}} &             \textbf{0.81} &                      3.75 &                      0.72 \\
           & TaskEmb-T &                      0.87 &                      2.13 &                      0.33 &             \textbf{0.72} &             \textbf{4.19} & \underline{\textbf{0.93}} &             \textbf{0.77} &             \textbf{3.61} & \underline{\textbf{0.00}} &                      0.80 &             \textbf{1.36} & \underline{\textbf{0.50}} &                      0.80 &             \textbf{2.82} &             \textbf{0.46} \\
\bottomrule
\end{tabular}
    }
    \caption{ BERT}
    \end{subtable}
    \vspace{-0.8em}
        \caption{Evaluation of intermediate task rankings produced by different methods for RoBERTa (a) and BERT (b) when preferring tasks of the same type.
    The table shows the mean \textit{NDCG} and \textit{Regret} scores by target task type.
    The best score in each group is highlighted in bold, best overall score is underlined.
    For \textit{NDCG}, higher is better; for \textit{Regret}, lower is better.}
    \label{tab:method_results_task_type}
    \vspace{-1em}
\end{table*}

\begin{table*}[]
    \centering
    \def\arraystretch{0.87}
    \resizebox{\linewidth}{!}{
    \begin{tabular}{llrrrrrrrrrrrr|rrr}
\toprule
& {} & \multicolumn{3}{c}{\textbf{Classification}} & \multicolumn{3}{c}{\textbf{M. Choice}} & \multicolumn{3}{c}{\textbf{QA}} & \multicolumn{3}{c}{\textbf{Tagging}}  & \multicolumn{3}{c}{\textbf{All}} \\
  \cmidrule(lr){3-5} \cmidrule(lr){6-8} \cmidrule(lr){9-11} \cmidrule(lr){12-14} \cmidrule(lr){15-17} 
 & {} &           NDCG &           R@1 &           R@3 &          NDCG &           R@1 &           R@3 &          NDCG &            R@1 &           R@3 &          NDCG &           R@1 &           R@3 &          NDCG &           R@1 &           R@3 \\
\cmidrule(lr){3-17}
\multirow{2}{*}{$D_S$} & Size+SEmb &                      0.72 &                      5.68 &             \textbf{0.47} &             \textbf{0.64} &             \textbf{6.95} &             \textbf{2.39} & \underline{\textbf{0.61}} &             \textbf{33.18} & \underline{\textbf{0.00}} &                      0.60 &             \textbf{7.81} &             \textbf{1.33} &             \textbf{0.66} &                     11.41 & \underline{\textbf{1.06}} \\
           & Size+TextEmb &             \textbf{0.77} &             \textbf{2.13} &                      1.13 &                      0.59 &                      6.98 &                      4.86 &                      0.52 &             \textbf{33.18} &                      2.05 &             \textbf{0.62} &                      7.94 &             \textbf{1.33} &                      0.65 &            \textbf{10.15} &                      2.35 \\
\cmidrule(lr){3-17}
\multirow{4}{*}{$M_S$} & FSFT+kNN+linear &             \textbf{0.91} & \underline{\textbf{0.19}} &             \textbf{0.12} &             \textbf{0.83} &             \textbf{1.91} & \underline{\textbf{0.00}} &                         - &                          - &                         - & \underline{\textbf{0.95}} & \underline{\textbf{0.11}} &             \textbf{0.11} &                         - &                         - &                         - \\
           & Size+FSFT &                      0.80 &                      1.21 &                      0.19 &                      0.55 &                      9.56 &                      2.39 &             \textbf{0.28} &             \textbf{57.20} &            \textbf{18.52} &                      0.66 &                      5.33 &                      0.47 &             \textbf{0.62} &            \textbf{14.42} &             \textbf{4.17} \\
           & Size+kNN &                      0.73 &                      6.42 &             \textbf{0.12} &                      0.50 &                      7.07 &                      4.30 &                         - &                          - &                         - &                      0.62 &                      8.44 &                      1.03 &                         - &                         - &                         - \\
           & Size+linear &                      0.70 &                      3.53 &                      2.44 &                      0.61 &                      4.30 &                      2.39 &                         - &                          - &                         - &                      0.66 &                      4.37 &                      0.47 &                         - &                         - &                         - \\
\cmidrule(lr){3-17}
\multirow{10}{*}{$D_S, M_S$} & FSFT+FS-TaskEmb &                      0.90 &                      0.46 & \underline{\textbf{0.00}} & \underline{\textbf{0.88}} &                      0.83 & \underline{\textbf{0.00}} &                      0.25 &                      34.25 &                     18.20 &             \textbf{0.93} &             \textbf{0.19} &                      0.19 & \underline{\textbf{0.78}} &                      6.66 &                      3.34 \\
           & SEmb+TaskEmb & \underline{\textbf{0.92}} &                      0.27 &                      0.27 &                      0.78 &                      4.88 & \underline{\textbf{0.00}} &                      0.30 &                      28.20 &                     20.15 &                      0.81 &                      0.65 &                      0.19 &                      0.75 &                      6.68 &                      3.80 \\
           & Size+FS-TaskEmb &                      0.83 &                      0.93 &                      0.17 &                      0.61 &                      9.87 &                      1.22 &                      0.31 & \underline{\textbf{16.39}} &                     12.90 &                      0.69 &                      1.54 &                      0.47 &                      0.65 &                      6.29 &                      2.83 \\
           & Size+SEmb+FSFT+FS-TaskEmb &                      0.91 & \underline{\textbf{0.19}} &                      0.19 &                      0.80 &                      4.88 & \underline{\textbf{0.00}} &             \textbf{0.33} &                      21.21 &            \textbf{10.38} &                      0.86 &                      0.65 &                      0.47 &                      0.76 & \underline{\textbf{5.38}} &             \textbf{2.04} \\
           & Size+SEmb+linear+TaskEmb &                      0.85 &                      3.45 &                      0.19 &                      0.79 &                      4.88 & \underline{\textbf{0.00}} &                         - &                          - &                         - &                      0.78 &                      1.54 &                      0.47 &                         - &                         - &                         - \\
           & Size+TaskEmb &                      0.66 &                      5.98 &                      1.15 &                      0.50 &                      9.56 &                      3.37 &                      0.28 &                      43.50 &                     32.70 &                      0.55 &                     64.25 &                      1.53 &                      0.53 &                     24.38 &                      7.56 \\
           & Size+TaskEmb+TextEmb &                      0.81 &                      2.13 &                      0.12 &                      0.67 &                      4.88 &                      2.39 &                      0.32 &                      38.07 &                     32.68 &                      0.68 &                     31.84 &                      0.47 &                      0.66 &                     14.82 &                      6.72 \\
           & TaskEmb+FS-TaskEmb &                      0.79 &                      5.53 &                      0.19 &                      0.71 &                      3.74 &                      1.22 &                      0.25 &                      22.40 &                     22.40 &                      0.85 &             \textbf{0.19} &                      0.19 &                      0.68 &                      7.14 &                      4.51 \\
           & TaskEmb+TextEmb &                      0.86 &                      1.12 &                      0.12 &                      0.76 &                      4.88 & \underline{\textbf{0.00}} &                      0.30 &                      36.36 &                     28.75 &                      0.83 &             \textbf{0.19} &                      0.19 &                      0.73 &                      8.39 &                      5.30 \\
           & All &                      0.90 & \underline{\textbf{0.19}} &                      0.19 &                      0.87 & \underline{\textbf{0.38}} & \underline{\textbf{0.00}} &                         - &                          - &                         - &                      0.91 &                      0.65 & \underline{\textbf{0.00}} &                         - &                         - &                         - \\
\bottomrule
\end{tabular}
    }
    \caption{Evaluation of intermediate task rankings produced by method combinations for RoBERTa.
    The table shows the mean \textit{NDCG} and \textit{Regret} scores by target task type.
    The best score in each group is highlighted in bold, best overall score is underlined.
    For \textit{NDCG}, higher is better; for \textit{Regret}, lower is better.}
    \label{tab:method_combinations}
\end{table*}

\begin{table*}[]
    \centering
    \def\arraystretch{0.87}
    \resizebox{\linewidth}{!}{

\begin{tabular}{lrrrrrrrrrrrr|rrr}
\toprule
& \multicolumn{3}{c}{\textbf{Classification}} & \multicolumn{3}{c}{\textbf{M. Choice}} & \multicolumn{3}{c}{\textbf{QA}} & \multicolumn{3}{c}{\textbf{Tagging}}  & \multicolumn{3}{c}{\textbf{All}} \\
  \cmidrule(lr){2-4} \cmidrule(lr){5-7} \cmidrule(lr){8-10} \cmidrule(lr){11-13} \cmidrule(lr){14-16} 
&           NDCG &           R@1 &           R@3 &          NDCG &           R@1 &           R@3 &          NDCG &            R@1 &           R@3 &          NDCG &           R@1 &           R@3 &          NDCG &           R@1 &           R@3 \\
\cmidrule(lr){2-16}
SEmb-BERT$_D$    &           0.83 & \textbf{0.27} & \textbf{0.19} &          0.67 &          8.13 &          2.56 &          0.48 &         17.04 &          7.16 &          0.84 & \textbf{0.47} & \textbf{0.00} &          0.72 &          5.50 &          2.07 \\
SEmb-BERT$_B$    &           0.82 & \textbf{0.27} &          0.24 &          0.58 &         12.85 &          2.56 & \textbf{0.61} & \textbf{7.16} & \textbf{2.05} &          0.84 & \textbf{0.47} & \textbf{0.00} &          0.72 &          4.99 &          1.16 \\

SEmb-BERT$_L$    &           0.82 & \textbf{0.27} &          0.27 &          0.58 &         11.07 &          2.56 &          0.49 &         17.04 &          7.16 &          0.84 & \textbf{0.47} &          0.11 &          0.70 &          6.30 &          2.12 \\
SEmb-RoBERTa$_D$ &  \textbf{0.84} & \textbf{0.27} & \textbf{0.19} &          0.76 &          4.86 & \textbf{0.00} &          0.52 &         17.04 &          4.04 & \textbf{0.88} & \textbf{0.47} &          0.11 & \textbf{0.77} &          4.60 &          0.82 \\
SEmb-RoBERTa$_B$ &           0.82 & \textbf{0.27} &          0.27 & \textbf{0.79} & \textbf{4.47} & \textbf{0.00} &          0.49 &         17.04 &          7.59 &          0.80 & \textbf{0.47} &          0.47 &          0.75 &          4.50 &          1.56 \\
SEmb-RoBERTa$_L$ &           0.76 &          3.73 &          0.27 &          0.75 & \textbf{4.47} & \textbf{0.00} &          0.59 & \textbf{7.16} & \textbf{2.05} &          0.82 & \textbf{0.47} & \textbf{0.00} &          0.74 & \textbf{3.96} & \textbf{0.47} \\
\bottomrule
\end{tabular}
    }
    \caption{Evaluation of intermediate task rankings produced by SEmb variations for RoBERTa tasks. \textit{D, B,} and \textit{L} stand for \textit{Distill}, \textit{Base}, and \textit{Large}, respectively.
    The table shows the mean \textit{NDCG} and \textit{Regret} scores by target task type.
    The best overall scores are highlighted in bold.
    For \textit{NDCG}, higher is better; for \textit{Regret}, lower is better. 
    }
    \label{tab:SEmb_types}
\end{table*}

\section{Rank Fusion}
\label{subsec:efficient_method_combinations} 
\citet{vuExploringPredictingTransferability2020} use the Reciprocal Rank Fusion algorithm \citep{cormackReciprocalRankFusion2009} 
to aggregate the rankings of TextEmb and TaskEmb. 
further experiment with various combinations of ranks produced by methods of different categories, e.g. \mbox{\textit{Size + SEmb}}.  
Table~\ref{tab:method_combinations} shows the results for a selection of all possible method combinations when applied to intermediate task selection for RoBERTa.

In a few cases, fusing improves performance over the single-method performances of all included methods (e.g. \textit{TaskEmb+TextEmb}).
However, for most cases, rank fusion performance is either roughly on-par with the performance of the best included single method (e.g. \textit{SEmb+TaskEmb}) or even hurts task selection performance sometimes significantly (e.g. \textit{Size+SEmb}).
Thus, while adding additional computational overhead to the task selection process, fusing does not yield better performance in general.

\section{SEmb Model Dependency} 

The full results of our experiments with sentence-embedding model variants can be found in  Table~\ref{tab:SEmb_types}. Experiments were conducted on RoBERTa transfer results.

\onecolumn


{\small
\begin{longtable}[c]{p{3cm} c p{2.5cm} p{2.3cm} p{1.5cm} p{1.4cm} p{1.4cm}}
    \toprule
    \textbf{Name} & |\textbf{Train}| & \textbf{Task} & \textbf{Domain/ Source} & \textbf{Metric(s)} & \textbf{RoBERTa} & \textbf{BERT} \\
    \midrule
    \textit{Sequence classification/ regression} \\
    \midrule
    MRPC \citep{dolanAutomaticallyConstructingCorpus2005} & 3.7K & semantic textual similarity & news & acc./ F1 & 88.48/ 91.53 & 84.80/ 89.53\\
    SICK \citep{marelliSICKCureEvaluation2014} & 4.4K & NLI & image/ video captions & acc. & 89.29 & 84.24 \\
    WiC \citep{pilehvarWiCWordinContextDataset2019} & 5.4K & word sense disambiguation & misc. & acc. & 65.52 & 65.99 \\
    TREC \citep{liLearningQuestionClassifiers2002} & 5.5K & question classification & misc. & acc. & 96.4 & 95.60 \\
    SciCite \citep{cohanStructuralScaffoldsCitation2019} & 8.2K & citation intents & scientific papers & acc. & 84.72 & 85.26 \\
    CoLA \citep{warstadtNeuralNetworkAcceptability2019} & 8.5K & linguistic acceptability & books, journals & Matthews & 59.18 & 62.18 \\
    Emotion \citep{saraviaCARERContextualizedAffect2018} & 16K & emotion classification & Twitter & acc. & 94.1 & 93.5 \\
    IMDb \citep{maasLearningWordVectors2011} & 25K & sentiment classification & movie reviews & acc. & 94.19 & 91.76 \\
    MultiRC \citep{khashabiLookingSurfaceChallenge2018} & 27K & reading comprehension  & misc. & EM/ F1 & 28.96/ 67.01 & 18.57/ 66.35 \\
    SciTail \citep{khotSciTaiLTextualEntailment2018} & 27K & NLI & science exams & acc. & 95.25 & 93.79 \\
    EmoContext \citep{chatterjeeSemEval2019TaskEmoContext2019} & 30K & emotion classification & crowdsourced & acc. & 89 & 89.74 \\
    SST-2 \citep{socherRecursiveDeepModels2013} & 67K & sentiment classification & movie reviews & acc. & 94.95 & 92.20 \\
    ReCoRD \citep{zhangReCoRDBridgingGap2018} & 101K & commonsense reasoning & news articles & EM/ F1 & 80.55/ 81.25 & 64.58/ 65.24 \\
    QNLI \citep{wangGLUEMultiTaskBenchmark2018} & 105K & question-answer NLI & Wikipedia & acc. & 92.75 & 91.14 \\
    ANLI \citep{nieAdversarialNLINew2020} & 163K & NLI & misc. & acc. & 41.5 & 45.42 \\
    QQP \citep{iyerFirstQuoraDataset2017} & 364K & semantic textual similarity & Quora & acc./ F1 & 90.80/ 87.68 & 90.31/ 87.04 \\
    MNLI \citep{williamsBroadCoverageChallengeCorpus2018} & 393K & NLI & misc. & acc. (matched) & 87.5 & 84.20 \\
    SNLI \citep{bowmanLargeAnnotatedCorpus2015} & 550K & NLI & misc. & acc. & 91.13 & 90.62 \\
    Yelp Polarity \citep{zhangCharacterlevelConvolutionalNetworks2015} & 560K & sentiment classification & Yelp reviews & acc. & 96.61 & 95.71 \\
    \midrule
    \textit{Multiple-choice} \\
    \midrule
    QuaRTz \citep{tafjordQuaRTzOpenDomainDataset2019} & 2.7K & qualitative reasoning & crowdsourced & acc. & 79.69 & 52.86 \\
    Cosmos QA \citep{huangCosmosQAMachine2019} & 25K & commonsense reasoning & crowdsourced & acc. & 70.49 & 60.47 \\
    Social IQA \citep{sapSocialIQACommonsense2019} & 33K & commonsense reasoning & knowledge base & acc. & 72.21 & 62.49 \\
    HellaSwag \citep{zellersHellaSwagCanMachine2019} & 40K & commonsense reasoning & misc. & acc. & 62.04 & 38.20 \\
    WinoGrande \citep{sakaguchiWINOGRANDEAdversarialWinograd2020} & 41K & coreference resolution & crowdsourced & acc. & 63.54 & 54.38 \\
    SWAG \citep{zellersSWAGLargeScaleAdversarial2018} & 74K & commonsense reasoning & video captions & acc. & 83.29 & 80.06 \\
    RACE \citep{laiRACELargescaleReAding2017} & 88K & reading comprehension & English exams & acc. & 73.46 & 65.97 \\
    ART \citep{bhagavatulaAbductiveCommonsenseReasoning2019} & 170K & NLI & stories & acc. & 73.43 & 64.36 \\
    \midrule
    \textit{Extractive question answering} \\
    \midrule
    Quoref \citep{dasigiQuorefReadingComprehension2019} & 20K & coreference QA & Wikipedia & EM/ F1 & 68.73/ 73.22 & 64.06/ 68.15\\
    WikiHop \citep{welblConstructingDatasetsMultihop2018}\footnotemark  & 51K & multi-hop QA & Wikipedia & EM/ F1 & 56.48/ 61.71 & 55.72/ 60.79 \\
    DuoRC-s \citep{sahaDuoRCComplexLanguage2018}\footnotemark[\value{footnote}]  & 86K & QA & Wikipedia & EM/ F1 & 59.36/ 67.10 & 53.19/ 60.73 \\
    HotpotQA \citep{yangHotpotQADatasetDiverse2018}\footnotemark[\value{footnote}]  & 90K & multi-hop QA & Wikipedia & EM/ F1 & 57.60/ 71.05 & 54.81/ 68.49 \\
    DuoRC-p \citep{sahaDuoRCComplexLanguage2018}\footnotemark[\value{footnote}]  & 100K & QA & IMDb & EM/ F1 & 49.76/ 53.38 & 47.76/ 51.31 \\
    SQuAD 1.0 \citep{rajpurkarSQuAD1000002016}\footnotemark[\value{footnote}]  & 108K & QA & Wikipedia & EM/ F1 & 84.02/ 91.06 & 80.26/ 88.08 \\
    NewsQA \citep{trischlerNewsQAMachineComprehension2017}\footnotemark[\value{footnote}]  & 120K & QA & news articles & EM/ F1 & 48.70/ 63.93 & 48.68/ 64.86 \\
    SQuAD 2.0 \citep{rajpurkarKnowWhatYou2018}\footnotemark[\value{footnote}]  & 162K & QA & Wikipedia & EM/ F1 & 78.39/ 81.47 & 67.99/ 71.22 \\
    \midrule
    \textit{Sequence tagging} \\
    \midrule
    NER-WNUT17 \citep{derczynskiResultsWNUT2017Shared2017} & 3.4K & NER & Twitter, forums & F1 & 55.24 & 45.27 \\
    NER-MITMovie & 7.8K & NER & movie reviews & F1 & 69.29 & 68.63 \\
    Chunk-CoNLL2000 \citep{sangIntroductionCoNLL2000Shared2000} & 8.9K & chunking & Penn Treebank & F1 & 96.35 & 95.92 \\
    POS-EWT \citep{silveiraGoldStandardDependency2014} & 12.5K & POS & web treebank & F1 & 97.30 & 96.79 \\
    POS-CoNLL2003 \citep{sangIntroductionCoNLL2003Shared2003} & 14K & POS & news & F1 & 95.05 & 93.96 \\
    GED-FCE \citep{reiCompositionalSequenceLabeling2016} & 29K & GED & misc. & $\text{F}_{0.5}$ & 89.79/ 68.12 & 64.94 \\
    ST-PMB \citep{abzianidzeUniversalSemanticTagging2017} & 63K & semantic tagging & meaning bank & acc./ F1 & 89.50/ 89.38 & 90.26/ 90.26 \\
    \bottomrule
    \caption{Overview of intermediate tasks used in our experiments, grouped by task type and ordered by training set size.}
    \label{table:source_tasks}
\end{longtable}
}

\footnotetext{We use the version provided in MultiQA \cite{talmorMultiQAEmpiricalInvestigation2019}.}

\begin{table*}[h!]
    \centering
    {\small 
    \begin{tabular}[t]{l l l l}
        \toprule
        \textbf{Name} & \textbf{Task} & \textbf{Domain/ Source} & \textbf{Metric(s)} \\
        \midrule
        \textit{Sequence classification/ regression} \\
        \midrule
        BoolQ \citep{clarkBoolQExploringSurprising2019} & binary QA & Wikipedia, web queries & acc. \\
        RTE \citep{daganPASCALRecognisingTextual2005} & NLI & news, Wikipedia & acc. \\
        Rotten Tomatoes \citep{pangSeeingStarsExploiting2005} & sentiment classification & movie reviews & acc. \\
        STS-B \citep{cerSemEval2017TaskSemantic2017} & semantic textual similarity & misc. & Spearman \\
        \midrule
        \textit{Multiple-choice} \\
        \midrule
        COPA \citep{gordonSemEval2012TaskChoice2012} & commonsense reasoning & blogs, encyclopedia & acc. \\
        CS QA \citep{talmorCommonsenseQAQuestionAnswering2019} & commonsense reasoning & knowledge base & acc. \\
        QuAIL \citep{rogersGettingCloserAI2020} & multiple-choice QA & misc. & acc. \\
        \midrule
        \textit{Extractive question answering} \\
        \midrule
        CQ \citep{baoConstraintBasedQuestionAnswering2016}\footnotemark[\value{footnote}] & QA & web snippets & EM/ F1 \\
        DROP \citep{duaDROPReadingComprehension2019}\footnotemark[\value{footnote}] & QA & Wikipedia & EM/ F1 \\
        \midrule
        \textit{Sequence labeling} \\
        \midrule
        DepRel-EWT \citep{silveiraGoldStandardDependency2014} & relation classification\footnotemark & web treebank & F1 \\
        NER-CoNLL2003 \citep{sangIntroductionCoNLL2003Shared2003} & NER & news & F1 \\
        \bottomrule
    \end{tabular}
    }
    \caption{Overview of target tasks used in our experiments, grouped by task type.}
    \label{table:target_tasks}
\end{table*}

\footnotetext{Instead of performing full dependency parsing, we only label each token in a sentence with a label corresponding to the dependency relation to its head as this task can be modeled directly as a sequence tagging task.}

\begin{table*}[t!]
    \centering
    \footnotesize
    \resizebox{\textwidth}{!}{
    \begin{tabular}{lp{1cm}p{1cm}p{1cm}p{1cm}p{1cm}p{1cm}p{1cm}p{1cm}p{1cm}p{1cm}p{1cm}}
    \toprule
    \bf Task & \bf BoolQ & \bf COPA & \bf CQ & \bf CS QA & \bf CoNLL 2003 & \bf DROP & \bf DepRel-EWT & \bf Quail & \bf R. Tomatoes & \bf RTE & \bf STS-B \\
    \midrule
No Transfer & \textbf{63.60} & \textbf{67.04} & \textbf{23.98} & \textbf{51.06} & \textbf{83.66} & \textbf{14.36} & \textbf{76.20} & \textbf{54.31} & \textbf{85.35} & \textbf{60.94} & \textbf{86.52} \\
\midrule
    Avg. Transfer & \textbf{67.29} & \textbf{67.99} & \textbf{25.76} & \textbf{50.28} & \textbf{82.33} & \textbf{13.96} & \textbf{75.06} & \textbf{54.83} & \textbf{85.16} & \textbf{67.21} & \textbf{86.72} \\
    \midrule
ANLI & \cellcolor[rgb]{ .412,  .757,  .502}75.84 & \cellcolor[rgb]{ .647,  .855,  .702}71.87 & \cellcolor[rgb]{ .918,  .965,  .929}26.55 & \cellcolor[rgb]{ .808,  .922,  .835}51.73 & \cellcolor[rgb]{ .961,  .984,  .969}83.80 & \cellcolor[rgb]{ 1,  .976,  .91}13.99 & \cellcolor[rgb]{ 1,  .973,  .89}74.66 & \cellcolor[rgb]{ .557,  .816,  .627}59.23 & \cellcolor[rgb]{ 1,  .882,  .529}84.08 & \cellcolor[rgb]{ .416,  .757,  .506}77.38 & \cellcolor[rgb]{ .678,  .867,  .729}87.77 \\
ART  & \cellcolor[rgb]{ .761,  .902,  .8}68.55 & \cellcolor[rgb]{ .694,  .875,  .741}71.27 & \cellcolor[rgb]{ 1,  .984,  .937}23.43 & \cellcolor[rgb]{ 1,  .996,  .996}51.05 & \cellcolor[rgb]{ 1,  .961,  .847}81.21 & \cellcolor[rgb]{ 1,  .906,  .631}12.85 & \cellcolor[rgb]{ 1,  .949,  .8}73.36 & \cellcolor[rgb]{ .867,  .945,  .886}55.82 & \cellcolor[rgb]{ 1,  .965,  .867}84.99 & \cellcolor[rgb]{ .698,  .875,  .745}69.43 & \cellcolor[rgb]{ .843,  .937,  .867}87.13 \\
CoLA & \cellcolor[rgb]{ .937,  .976,  .945}64.96 & \cellcolor[rgb]{ .863,  .945,  .886}68.93 & \cellcolor[rgb]{ .98,  .992,  .984}24.61 & \cellcolor[rgb]{ .855,  .941,  .878}51.57 & \cellcolor[rgb]{ 1,  .976,  .91}82.25 & \cellcolor[rgb]{ 1,  .961,  .847}13.74 & \cellcolor[rgb]{ .894,  .957,  .91}77.02 & \cellcolor[rgb]{ .957,  .984,  .965}54.81 & \cellcolor[rgb]{ .976,  .992,  .98}85.62 & \cellcolor[rgb]{ .882,  .953,  .902}64.26 & \cellcolor[rgb]{ 1,  .953,  .82}86.24 \\
CoNLL'00 & \cellcolor[rgb]{ .984,  .992,  .984}64.00 & \cellcolor[rgb]{ 1,  .804,  .212}57.13 & \cellcolor[rgb]{ 1,  .957,  .827}22.49 & \cellcolor[rgb]{ 1,  .804,  .216}43.57 & \cellcolor[rgb]{ 1,  .976,  .91}82.23 & \cellcolor[rgb]{ 1,  .776,  .094}10.62 & \cellcolor[rgb]{ .635,  .851,  .694}78.97 & \cellcolor[rgb]{ 1,  .839,  .357}51.88 & \cellcolor[rgb]{ 1,  .965,  .867}84.99 & \cellcolor[rgb]{ .867,  .945,  .886}64.74 & \cellcolor[rgb]{ 1,  .753,  0}84.98 \\
Cosmos QA & \cellcolor[rgb]{ .627,  .847,  .686}71.36 & \cellcolor[rgb]{ .514,  .796,  .588}73.73 & \cellcolor[rgb]{ .969,  .988,  .973}25.00 & \cellcolor[rgb]{ .722,  .886,  .765}52.03 & \cellcolor[rgb]{ 1,  .976,  .914}82.29 & \cellcolor[rgb]{ 1,  .973,  .89}13.92 & \cellcolor[rgb]{ 1,  .98,  .929}75.21 & \cellcolor[rgb]{ .773,  .906,  .808}56.84 & \cellcolor[rgb]{ 1,  .992,  .98}85.30 & \cellcolor[rgb]{ .494,  .788,  .573}75.21 & \cellcolor[rgb]{ .898,  .957,  .914}86.92 \\
DuoRC-p & \cellcolor[rgb]{ .78,  .91,  .816}68.17 & \cellcolor[rgb]{ 1,  .988,  .957}66.53 & \cellcolor[rgb]{ .569,  .824,  .635}36.97 & \cellcolor[rgb]{ .792,  .914,  .824}51.79 & \cellcolor[rgb]{ .467,  .776,  .549}85.47 & \cellcolor[rgb]{ .631,  .847,  .686}17.22 & \cellcolor[rgb]{ 1,  .976,  .918}75.03 & \cellcolor[rgb]{ 1,  .898,  .588}52.76 & \cellcolor[rgb]{ 1,  .973,  .89}85.05 & \cellcolor[rgb]{ .765,  .902,  .8}67.63 & \cellcolor[rgb]{ .651,  .855,  .706}87.88 \\
DuoRC-s & \cellcolor[rgb]{ .769,  .906,  .804}68.43 & \cellcolor[rgb]{ .725,  .886,  .769}70.80 & \cellcolor[rgb]{ .435,  .765,  .522}40.97 & \cellcolor[rgb]{ .651,  .855,  .702}52.28 & \cellcolor[rgb]{ .4,  .753,  .494}85.69 & \cellcolor[rgb]{ .643,  .851,  .698}17.13 & \cellcolor[rgb]{ .969,  .988,  .973}76.46 & \cellcolor[rgb]{ 1,  .965,  .867}53.80 & \cellcolor[rgb]{ 1,  .984,  .949}85.21 & \cellcolor[rgb]{ .82,  .925,  .847}66.06 & \cellcolor[rgb]{ .827,  .929,  .855}87.20 \\
EmoContext & \cellcolor[rgb]{ .847,  .937,  .871}66.79 & \cellcolor[rgb]{ .863,  .945,  .886}68.93 & \cellcolor[rgb]{ 1,  .965,  .867}22.84 & \cellcolor[rgb]{ 1,  .98,  .933}50.42 & \cellcolor[rgb]{ 1,  .976,  .91}82.23 & \cellcolor[rgb]{ 1,  .969,  .875}13.85 & \cellcolor[rgb]{ .914,  .965,  .925}76.88 & \cellcolor[rgb]{ .992,  .996,  .992}54.42 & \cellcolor[rgb]{ .945,  .98,  .957}85.93 & \cellcolor[rgb]{ .827,  .929,  .855}65.82 & \cellcolor[rgb]{ 1,  .988,  .953}86.44 \\
Emotion & \cellcolor[rgb]{ .898,  .957,  .914}65.79 & \cellcolor[rgb]{ 1,  .98,  .925}66.13 & \cellcolor[rgb]{ 1,  .898,  .588}20.42 & \cellcolor[rgb]{ 1,  .976,  .902}50.15 & \cellcolor[rgb]{ 1,  .98,  .922}82.38 & \cellcolor[rgb]{ 1,  .957,  .824}13.65 & \cellcolor[rgb]{ 1,  .953,  .808}73.49 & \cellcolor[rgb]{ .965,  .988,  .973}54.71 & \cellcolor[rgb]{ 1,  .941,  .773}84.74 & \cellcolor[rgb]{ .89,  .957,  .91}64.02 & \cellcolor[rgb]{ 1,  .949,  .804}86.22 \\
GED-FCE & \cellcolor[rgb]{ .976,  .992,  .98}64.11 & \cellcolor[rgb]{ .898,  .957,  .914}68.47 & \cellcolor[rgb]{ 1,  .953,  .816}22.37 & \cellcolor[rgb]{ 1,  .953,  .82}49.36 & \cellcolor[rgb]{ 1,  .98,  .925}82.47 & \cellcolor[rgb]{ 1,  .878,  .518}12.37 & \cellcolor[rgb]{ .784,  .914,  .82}77.84 & \cellcolor[rgb]{ .996,  1,  .996}54.36 & \cellcolor[rgb]{ .98,  .992,  .984}85.55 & \cellcolor[rgb]{ .914,  .965,  .925}63.42 & \cellcolor[rgb]{ 1,  .937,  .757}86.14 \\
Hellaswag & \cellcolor[rgb]{ .894,  .957,  .91}65.81 & \cellcolor[rgb]{ .863,  .945,  .886}68.93 & \cellcolor[rgb]{ 1,  .906,  .627}20.74 & \cellcolor[rgb]{ .902,  .961,  .918}51.41 & \cellcolor[rgb]{ 1,  .969,  .875}81.65 & \cellcolor[rgb]{ 1,  .949,  .804}13.56 & \cellcolor[rgb]{ 1,  .992,  .969}75.78 & \cellcolor[rgb]{ .871,  .949,  .894}55.75 & \cellcolor[rgb]{ 1,  .961,  .843}84.93 & \cellcolor[rgb]{ .749,  .898,  .788}67.99 & \cellcolor[rgb]{ .957,  .984,  .965}86.69 \\
HotpotQA & \cellcolor[rgb]{ .929,  .973,  .937}65.14 & \cellcolor[rgb]{ 1,  .965,  .867}65.40 & \cellcolor[rgb]{ .388,  .745,  .482}\textbf{42.35} & \cellcolor[rgb]{ 1,  .922,  .682}48.02 & \cellcolor[rgb]{ 1,  .957,  .835}80.99 & \cellcolor[rgb]{ .537,  .808,  .612}17.91 & \cellcolor[rgb]{ 1,  .867,  .463}68.55 & \cellcolor[rgb]{ 1,  .776,  .11}50.94 & \cellcolor[rgb]{ 1,  .851,  .404}83.74 & \cellcolor[rgb]{ .796,  .918,  .827}66.67 & \cellcolor[rgb]{ 1,  .89,  .553}85.84 \\
IMDb & \cellcolor[rgb]{ .812,  .922,  .839}67.55 & \cellcolor[rgb]{ .804,  .922,  .835}69.73 & \cellcolor[rgb]{ 1,  .984,  .949}23.56 & \cellcolor[rgb]{ 1,  .984,  .945}50.56 & \cellcolor[rgb]{ 1,  .98,  .925}82.49 & \cellcolor[rgb]{ 1,  .98,  .925}14.06 & \cellcolor[rgb]{ 1,  .988,  .953}75.56 & \cellcolor[rgb]{ .961,  .984,  .969}54.74 & \cellcolor[rgb]{ .882,  .953,  .898}86.62 & \cellcolor[rgb]{ 1,  .965,  .867}60.65 & \cellcolor[rgb]{ 1,  .949,  .8}86.21 \\
MIT Movie & 63.64 & \cellcolor[rgb]{ 1,  .98,  .929}66.20 & \cellcolor[rgb]{ 1,  .988,  .965}23.69 & \cellcolor[rgb]{ 1,  .937,  .745}48.65 & \cellcolor[rgb]{ .725,  .886,  .769}84.60 & \cellcolor[rgb]{ 1,  .863,  .451}12.10 & \cellcolor[rgb]{ .953,  .98,  .961}76.58 & \cellcolor[rgb]{ 1,  .922,  .69}53.14 & \cellcolor[rgb]{ .941,  .976,  .949}85.99 & \cellcolor[rgb]{ .863,  .945,  .882}64.86 & \cellcolor[rgb]{ 1,  .859,  .439}85.66 \\
MNLI & \cellcolor[rgb]{ .388,  .745,  .482}\textbf{76.26} & \cellcolor[rgb]{ .859,  .941,  .882}69.00 & \cellcolor[rgb]{ .992,  .996,  .992}24.32 & \cellcolor[rgb]{ .627,  .847,  .682}52.36 & \cellcolor[rgb]{ 1,  .976,  .91}82.23 & \cellcolor[rgb]{ .988,  .996,  .992}14.45 & \cellcolor[rgb]{ 1,  .949,  .8}73.39 & \cellcolor[rgb]{ .518,  .8,  .592}59.69 & \cellcolor[rgb]{ .941,  .976,  .949}85.99 & \cellcolor[rgb]{ .388,  .745,  .482}\textbf{78.10} & \cellcolor[rgb]{ .388,  .745,  .482}\textbf{88.90} \\
MRPC & \cellcolor[rgb]{ .796,  .918,  .827}67.88 & \cellcolor[rgb]{ .584,  .827,  .651}72.73 & \cellcolor[rgb]{ 1,  .98,  .925}23.35 & \cellcolor[rgb]{ .627,  .847,  .682}52.36 & \cellcolor[rgb]{ 1,  .988,  .953}82.90 & \cellcolor[rgb]{ 1,  .988,  .965}14.23 & \cellcolor[rgb]{ 1,  .988,  .957}75.59 & \cellcolor[rgb]{ .98,  .992,  .984}54.54 & \cellcolor[rgb]{ .949,  .98,  .957}85.90 & \cellcolor[rgb]{ .788,  .914,  .824}66.91 & \cellcolor[rgb]{ .827,  .929,  .855}87.19 \\
MultiRC & \cellcolor[rgb]{ .686,  .871,  .733}70.14 & \cellcolor[rgb]{ .659,  .859,  .71}71.73 & \cellcolor[rgb]{ 1,  .949,  .796}22.20 & \cellcolor[rgb]{ .388,  .745,  .482}\textbf{53.18} & \cellcolor[rgb]{ .988,  .996,  .992}83.71 & \cellcolor[rgb]{ 1,  .98,  .922}14.05 & \cellcolor[rgb]{ 1,  .98,  .933}75.25 & \cellcolor[rgb]{ .776,  .906,  .812}56.82 & \cellcolor[rgb]{ .98,  .992,  .984}85.58 & \cellcolor[rgb]{ .549,  .812,  .62}73.65 & \cellcolor[rgb]{ .851,  .937,  .875}87.11 \\
NewsQA & \cellcolor[rgb]{ .757,  .898,  .792}68.71 & \cellcolor[rgb]{ .58,  .827,  .647}72.80 & \cellcolor[rgb]{ .447,  .773,  .533}40.60 & \cellcolor[rgb]{ 1,  .98,  .922}50.34 & \cellcolor[rgb]{ .933,  .973,  .945}83.89 & \cellcolor[rgb]{ .51,  .796,  .584}18.13 & \cellcolor[rgb]{ 1,  .965,  .859}74.18 & \cellcolor[rgb]{ 1,  .894,  .58}52.73 & \cellcolor[rgb]{ 1,  .953,  .808}84.83 & \cellcolor[rgb]{ .792,  .914,  .827}66.79 & \cellcolor[rgb]{ .675,  .867,  .725}87.80 \\
POS-Co.'03 & \cellcolor[rgb]{ 1,  .984,  .937}63.52 & \cellcolor[rgb]{ 1,  .8,  .196}56.93 & \cellcolor[rgb]{ 1,  .875,  .502}19.64 & \cellcolor[rgb]{ 1,  .839,  .353}44.88 & \cellcolor[rgb]{ .69,  .871,  .737}84.72 & \cellcolor[rgb]{ 1,  .824,  .298}11.46 & \cellcolor[rgb]{ .467,  .776,  .549}80.27 & \cellcolor[rgb]{ 1,  .839,  .349}51.85 & \cellcolor[rgb]{ 1,  .922,  .682}84.49 & 61.01 & \cellcolor[rgb]{ 1,  .792,  .173}85.25 \\
POS-EWT & \cellcolor[rgb]{ .961,  .984,  .965}64.48 & \cellcolor[rgb]{ 1,  .918,  .667}62.87 & \cellcolor[rgb]{ 1,  .969,  .878}22.93 & \cellcolor[rgb]{ 1,  .925,  .698}48.18 & \cellcolor[rgb]{ .647,  .855,  .702}84.86 & \cellcolor[rgb]{ 1,  .78,  .125}10.74 & \cellcolor[rgb]{ .388,  .745,  .482}\textbf{80.84} & \cellcolor[rgb]{ 1,  .922,  .694}53.16 & \cellcolor[rgb]{ 1,  .882,  .529}84.08 & \cellcolor[rgb]{ .882,  .953,  .902}64.26 & \cellcolor[rgb]{ 1,  .953,  .812}86.23 \\
QNLI & \cellcolor[rgb]{ .737,  .89,  .776}69.09 & \cellcolor[rgb]{ .659,  .859,  .71}71.73 & \cellcolor[rgb]{ .631,  .847,  .686}35.15 & \cellcolor[rgb]{ .525,  .804,  .596}52.72 & \cellcolor[rgb]{ 1,  .996,  .984}83.47 & \cellcolor[rgb]{ .816,  .925,  .843}15.79 & \cellcolor[rgb]{ .929,  .973,  .941}76.76 & \cellcolor[rgb]{ .808,  .922,  .839}56.45 & \cellcolor[rgb]{ 1,  .91,  .647}84.40 & \cellcolor[rgb]{ .714,  .882,  .757}69.07 & \cellcolor[rgb]{ .6,  .835,  .663}88.08 \\
QQP  & \cellcolor[rgb]{ .745,  .894,  .784}68.88 & \cellcolor[rgb]{ .506,  .796,  .584}73.80 & \cellcolor[rgb]{ 1,  .969,  .878}22.94 & \cellcolor[rgb]{ 1,  .941,  .773}48.89 & \cellcolor[rgb]{ 1,  .965,  .855}81.34 & \cellcolor[rgb]{ 1,  .824,  .29}11.42 & \cellcolor[rgb]{ 1,  .937,  .749}72.63 & \cellcolor[rgb]{ 1,  .965,  .859}53.77 & \cellcolor[rgb]{ 1,  .988,  .957}85.24 & \cellcolor[rgb]{ .675,  .867,  .725}70.16 & \cellcolor[rgb]{ .616,  .843,  .675}88.02 \\
QuaRTz & \cellcolor[rgb]{ .976,  .992,  .98}64.14 & \cellcolor[rgb]{ 1,  .925,  .71}63.40 & \cellcolor[rgb]{ 1,  .965,  .867}22.84 & \cellcolor[rgb]{ .769,  .906,  .804}51.87 & \cellcolor[rgb]{ 1,  .996,  .984}83.45 & \cellcolor[rgb]{ 1,  .984,  .949}14.16 & \cellcolor[rgb]{ 1,  .996,  .992}76.13 & \cellcolor[rgb]{ .996,  1,  .996}54.37 & \cellcolor[rgb]{ .961,  .984,  .969}85.77 & \cellcolor[rgb]{ 1,  .992,  .976}60.89 & \cellcolor[rgb]{ 1,  .98,  .925}86.40 \\
Quoref & \cellcolor[rgb]{ .843,  .933,  .867}66.92 & \cellcolor[rgb]{ .388,  .745,  .482}\textbf{75.40} & \cellcolor[rgb]{ .651,  .855,  .706}34.47 & \cellcolor[rgb]{ 1,  .992,  .976}50.83 & \cellcolor[rgb]{ .475,  .78,  .553}85.45 & \cellcolor[rgb]{ .596,  .831,  .659}17.47 & \cellcolor[rgb]{ .773,  .906,  .808}77.94 & \cellcolor[rgb]{ .984,  .996,  .988}54.50 & \cellcolor[rgb]{ 1,  .929,  .725}84.62 & \cellcolor[rgb]{ .784,  .91,  .82}67.03 & \cellcolor[rgb]{ .761,  .902,  .796}87.46 \\
RACE & \cellcolor[rgb]{ .549,  .812,  .62}73.00 & \cellcolor[rgb]{ .58,  .827,  .647}72.80 & \cellcolor[rgb]{ 1,  .922,  .682}21.23 & \cellcolor[rgb]{ 1,  .969,  .882}49.93 & \cellcolor[rgb]{ 1,  .827,  .306}72.29 & \cellcolor[rgb]{ 1,  .91,  .643}12.90 & \cellcolor[rgb]{ 1,  .882,  .529}69.47 & \cellcolor[rgb]{ .388,  .745,  .482}\textbf{61.09} & \cellcolor[rgb]{ .953,  .98,  .961}85.87 & \cellcolor[rgb]{ .557,  .816,  .627}73.41 & \cellcolor[rgb]{ .545,  .812,  .616}88.29 \\
ReCoRD & \cellcolor[rgb]{ 1,  .753,  0}62.17 & \cellcolor[rgb]{ 1,  .882,  .529}61.13 & \cellcolor[rgb]{ .875,  .949,  .894}27.86 & \cellcolor[rgb]{ 1,  .976,  .914}50.26 & \cellcolor[rgb]{ 1,  .969,  .871}81.59 & \cellcolor[rgb]{ 1,  .898,  .596}12.69 & \cellcolor[rgb]{ 1,  .89,  .565}70.01 & \cellcolor[rgb]{ .835,  .933,  .859}56.16 & \cellcolor[rgb]{ 1,  .953,  .808}84.83 & \cellcolor[rgb]{ .831,  .929,  .859}65.70 & \cellcolor[rgb]{ 1,  .847,  .384}85.58 \\
SICK & \cellcolor[rgb]{ .824,  .925,  .851}67.32 & \cellcolor[rgb]{ .643,  .851,  .698}71.93 & \cellcolor[rgb]{ 1,  .918,  .678}21.18 & \cellcolor[rgb]{ .549,  .812,  .616}52.63 & \cellcolor[rgb]{ 1,  .988,  .957}82.99 & 14.37 & \cellcolor[rgb]{ 1,  .992,  .98}75.93 & \cellcolor[rgb]{ .906,  .961,  .922}55.36 & \cellcolor[rgb]{ 1,  .988,  .957}85.24 & \cellcolor[rgb]{ .765,  .902,  .8}67.63 & \cellcolor[rgb]{ .824,  .925,  .851}87.22 \\
SNLI & \cellcolor[rgb]{ .639,  .851,  .698}71.07 & \cellcolor[rgb]{ .875,  .949,  .894}68.80 & \cellcolor[rgb]{ 1,  .753,  0}15.26 & \cellcolor[rgb]{ 1,  .89,  .557}46.82 & \cellcolor[rgb]{ 1,  .753,  0}67.23 & \cellcolor[rgb]{ 1,  .753,  0}10.22 & \cellcolor[rgb]{ 1,  .753,  0}61.90 & \cellcolor[rgb]{ 1,  .957,  .831}53.68 & \cellcolor[rgb]{ 1,  .753,  0}82.65 & \cellcolor[rgb]{ .569,  .824,  .635}73.04 & \cellcolor[rgb]{ 1,  .812,  .243}85.36 \\
SQuAD & \cellcolor[rgb]{ .714,  .882,  .761}69.54 & \cellcolor[rgb]{ .898,  .957,  .914}68.47 & \cellcolor[rgb]{ .553,  .816,  .624}37.45 & \cellcolor[rgb]{ .612,  .839,  .671}52.42 & \cellcolor[rgb]{ .388,  .745,  .482}\textbf{85.73} & \cellcolor[rgb]{ .424,  .761,  .514}18.80 & \cellcolor[rgb]{ .863,  .945,  .886}77.24 & \cellcolor[rgb]{ 1,  .957,  .835}53.70 & \cellcolor[rgb]{ 1,  .961,  .855}84.96 & \cellcolor[rgb]{ .718,  .882,  .761}68.95 & \cellcolor[rgb]{ .843,  .937,  .867}87.13 \\
SQuAD 2.0 & \cellcolor[rgb]{ .682,  .871,  .733}70.18 & \cellcolor[rgb]{ .922,  .969,  .933}68.13 & \cellcolor[rgb]{ .49,  .788,  .569}39.30 & \cellcolor[rgb]{ .549,  .812,  .616}52.63 & \cellcolor[rgb]{ .455,  .773,  .537}85.51 & \cellcolor[rgb]{ .388,  .745,  .482}\textbf{19.05} & \cellcolor[rgb]{ .847,  .937,  .871}77.39 & \cellcolor[rgb]{ .953,  .98,  .961}54.87 & \cellcolor[rgb]{ .996,  1,  .996}85.40 & \cellcolor[rgb]{ .725,  .886,  .769}68.71 & \cellcolor[rgb]{ .788,  .914,  .824}87.34 \\
SST-2 & \cellcolor[rgb]{ .839,  .933,  .863}66.99 & \cellcolor[rgb]{ .808,  .922,  .839}69.67 & \cellcolor[rgb]{ 1,  .925,  .71}21.46 & \cellcolor[rgb]{ 1,  .98,  .922}50.31 & \cellcolor[rgb]{ 1,  .945,  .78}80.08 & \cellcolor[rgb]{ 1,  .878,  .518}12.37 & \cellcolor[rgb]{ 1,  .965,  .871}74.37 & \cellcolor[rgb]{ .886,  .953,  .902}55.61 & \cellcolor[rgb]{ .388,  .745,  .482}\textbf{91.78} & \cellcolor[rgb]{ .91,  .965,  .925}63.54 & \cellcolor[rgb]{ 1,  .973,  .894}86.36 \\
ST-PMB & \cellcolor[rgb]{ .957,  .984,  .965}64.51 & \cellcolor[rgb]{ 1,  .753,  0}54.47 & \cellcolor[rgb]{ 1,  .863,  .451}19.21 & \cellcolor[rgb]{ 1,  .753,  0}41.47 & \cellcolor[rgb]{ .655,  .859,  .706}84.84 & \cellcolor[rgb]{ 1,  .784,  .137}10.79 & \cellcolor[rgb]{ .573,  .824,  .639}79.46 & \cellcolor[rgb]{ 1,  .753,  0}50.52 & \cellcolor[rgb]{ 1,  .792,  .161}83.08 & \cellcolor[rgb]{ .961,  .984,  .969}62.09 & \cellcolor[rgb]{ 1,  .757,  .027}85.03 \\
SWAG & \cellcolor[rgb]{ .922,  .969,  .933}65.26 & \cellcolor[rgb]{ 1,  .984,  .941}66.33 & \cellcolor[rgb]{ 1,  .953,  .808}22.33 & \cellcolor[rgb]{ .549,  .812,  .616}52.63 & \cellcolor[rgb]{ 1,  .996,  .992}83.55 & \cellcolor[rgb]{ 1,  .973,  .894}13.93 & \cellcolor[rgb]{ .961,  .984,  .969}76.50 & \cellcolor[rgb]{ .812,  .922,  .843}56.41 & \cellcolor[rgb]{ 1,  .984,  .949}85.21 & \cellcolor[rgb]{ .631,  .847,  .686}71.36 & \cellcolor[rgb]{ .914,  .965,  .925}86.86 \\
SciCite & \cellcolor[rgb]{ .898,  .961,  .914}65.74 & \cellcolor[rgb]{ .937,  .976,  .945}67.93 & \cellcolor[rgb]{ 1,  .945,  .78}22.08 & \cellcolor[rgb]{ .988,  .996,  .988}51.11 & \cellcolor[rgb]{ .992,  1,  .996}83.69 & \cellcolor[rgb]{ 1,  .976,  .91}14.00 & \cellcolor[rgb]{ .945,  .976,  .953}76.63 & \cellcolor[rgb]{ .933,  .973,  .945}55.07 & \cellcolor[rgb]{ .992,  .996,  .992}85.46 & \cellcolor[rgb]{ 1,  .753,  0}58.72 & \cellcolor[rgb]{ .949,  .98,  .957}86.72 \\
SciTail & \cellcolor[rgb]{ .682,  .867,  .729}70.23 & \cellcolor[rgb]{ .639,  .851,  .694}72.00 & \cellcolor[rgb]{ 1,  .937,  .753}21.84 & \cellcolor[rgb]{ .565,  .82,  .631}52.58 & \cellcolor[rgb]{ 1,  .988,  .965}83.11 & \cellcolor[rgb]{ 1,  .988,  .965}14.23 & \cellcolor[rgb]{ .976,  .992,  .98}76.39 & \cellcolor[rgb]{ .851,  .937,  .875}55.99 & \cellcolor[rgb]{ .973,  .988,  .976}85.65 & \cellcolor[rgb]{ .6,  .835,  .663}72.20 & \cellcolor[rgb]{ .698,  .875,  .745}87.70 \\
Social IQA & \cellcolor[rgb]{ .663,  .863,  .718}70.60 & \cellcolor[rgb]{ .486,  .788,  .569}74.07 & \cellcolor[rgb]{ 1,  .933,  .741}21.74 & \cellcolor[rgb]{ .557,  .816,  .624}52.61 & \cellcolor[rgb]{ 1,  .922,  .682}78.50 & \cellcolor[rgb]{ 1,  .961,  .855}13.77 & \cellcolor[rgb]{ .949,  .98,  .957}76.61 & \cellcolor[rgb]{ .718,  .882,  .761}57.46 & \cellcolor[rgb]{ 1,  .961,  .855}84.96 & \cellcolor[rgb]{ .616,  .843,  .678}71.72 & \cellcolor[rgb]{ .875,  .949,  .894}87.02 \\
TREC & \cellcolor[rgb]{ .961,  .984,  .965}64.48 & \cellcolor[rgb]{ .961,  .984,  .969}67.60 & \cellcolor[rgb]{ 1,  .965,  .867}22.83 & \cellcolor[rgb]{ .722,  .886,  .765}52.03 & \cellcolor[rgb]{ 1,  .98,  .933}82.60 & \cellcolor[rgb]{ 1,  .98,  .929}14.09 & \cellcolor[rgb]{ .855,  .941,  .878}77.30 & \cellcolor[rgb]{ .898,  .961,  .914}55.45 & \cellcolor[rgb]{ .98,  .992,  .984}85.58 & \cellcolor[rgb]{ .871,  .949,  .89}64.62 & 86.53 \\
WNUT17 & \cellcolor[rgb]{ .973,  .988,  .976}64.20 & \cellcolor[rgb]{ 1,  .953,  .816}64.73 & \cellcolor[rgb]{ 1,  .953,  .824}22.45 & \cellcolor[rgb]{ 1,  .969,  .871}49.85 & \cellcolor[rgb]{ .757,  .902,  .796}84.49 & \cellcolor[rgb]{ 1,  .867,  .467}12.17 & \cellcolor[rgb]{ 1,  .992,  .969}75.78 & \cellcolor[rgb]{ 1,  .976,  .918}54.00 & \cellcolor[rgb]{ 1,  .953,  .808}84.83 & \cellcolor[rgb]{ .918,  .969,  .929}63.30 & \cellcolor[rgb]{ .996,  1,  .996}86.54 \\
WiC  & \cellcolor[rgb]{ .98,  .992,  .984}64.04 & \cellcolor[rgb]{ .627,  .847,  .686}72.13 & \cellcolor[rgb]{ 1,  .945,  .784}22.11 & \cellcolor[rgb]{ 1,  .98,  .922}50.31 & \cellcolor[rgb]{ 1,  .996,  .984}83.46 & \cellcolor[rgb]{ 1,  .984,  .937}14.11 & \cellcolor[rgb]{ 1,  .973,  .898}74.76 & \cellcolor[rgb]{ .918,  .965,  .929}55.25 & \cellcolor[rgb]{ 1,  .965,  .867}84.99 & \cellcolor[rgb]{ .945,  .976,  .953}62.58 & \cellcolor[rgb]{ .996,  1,  1}86.53 \\
WikiHop & \cellcolor[rgb]{ 1,  .886,  .541}62.95 & \cellcolor[rgb]{ 1,  .902,  .604}62.07 & \cellcolor[rgb]{ .502,  .792,  .576}39.02 & \cellcolor[rgb]{ 1,  .925,  .706}48.24 & \cellcolor[rgb]{ .855,  .941,  .878}84.16 & \cellcolor[rgb]{ .859,  .941,  .878}15.47 & \cellcolor[rgb]{ 1,  .91,  .643}71.11 & \cellcolor[rgb]{ 1,  .843,  .376}51.96 & \cellcolor[rgb]{ 1,  .902,  .612}84.30 & \cellcolor[rgb]{ .827,  .929,  .855}65.82 & \cellcolor[rgb]{ 1,  .827,  .314}85.47 \\
WinoGrande & \cellcolor[rgb]{ .792,  .914,  .824}67.92 & \cellcolor[rgb]{ 1,  .969,  .875}65.47 & \cellcolor[rgb]{ 1,  .894,  .576}20.31 & \cellcolor[rgb]{ 1,  .949,  .792}49.09 & \cellcolor[rgb]{ 1,  .973,  .894}81.93 & \cellcolor[rgb]{ 1,  .976,  .906}13.98 & \cellcolor[rgb]{ 1,  .953,  .82}73.65 & \cellcolor[rgb]{ .925,  .969,  .937}55.14 & \cellcolor[rgb]{ 1,  .784,  .137}83.02 & \cellcolor[rgb]{ .694,  .875,  .741}69.55 & \cellcolor[rgb]{ .902,  .961,  .918}86.91 \\
Yelp Polarity & \cellcolor[rgb]{ .882,  .953,  .902}66.04 & \cellcolor[rgb]{ 1,  .925,  .71}63.40 & \cellcolor[rgb]{ 1,  .878,  .518}19.79 & \cellcolor[rgb]{ 1,  .933,  .737}48.57 & \cellcolor[rgb]{ 1,  .89,  .565}76.51 & \cellcolor[rgb]{ 1,  .761,  .043}10.40 & \cellcolor[rgb]{ 1,  .886,  .541}69.65 & \cellcolor[rgb]{ 1,  .988,  .961}54.16 & \cellcolor[rgb]{ 1,  .984,  .937}85.18 & \cellcolor[rgb]{ .898,  .957,  .914}63.90 & \cellcolor[rgb]{ 1,  .878,  .506}85.76 \\
    \bottomrule
    \end{tabular}
    }
    \caption{Target task performances for transferring between intermediate tasks (rows) and target tasks (columns) with BERT as base model. The first row `\textit{No Transfer}' shows the baseline performance when training only on the target task without transfer. All scores are mean values over three random restarts.}
    \label{tab:transfer_results_BERT}
\end{table*}

\end{document}